\documentclass[preprint,7pt]{elsarticle}

\usepackage{xcolor}

\usepackage{svg}
\usepackage{amsmath}
\usepackage{amssymb}

\usepackage{subcaption}
\usepackage{mathtools}
\usepackage{graphicx}
\usepackage{color}
\usepackage{changepage}
\usepackage{adjustbox}
\usepackage[colorlinks,citecolor=red,urlcolor=blue,bookmarks=false,hypertexnames=true]{hyperref}
\usepackage{float}
\usepackage{multirow}
\usepackage[textsize=footnotesize]{todonotes}

\journal{Information Sciences}

\begin{document}

\begin{frontmatter}



\title{Interpretable Spectral Variational AutoEncoder (ISVAE) for time series clustering}


\author[uc3m,grg,eb2]{Óscar Jiménez Rama }
\author[oxford]{Fernando Moreno-Pino }
\author[uc3m,grg]{David Ramírez }
\author[uc3m,grg]{Pablo M. Olmos }

\address[uc3m]{Signal Processing and Learning Group, Universidad Carlos III de Madrid, Spain}
\address[grg]{Instituto de Investigación Sanitaria Gregorio Marañón, Madrid, Spain}
\address[eb2]{Evidence Based Behaviour (eB2), Madrid, Spain}
\address[oxford]{Oxford-Man Institute of Quantitative Finance, University of Oxford, UK}

\begin{abstract}


The best encoding is the one that is interpretable in nature. In this work, we introduce a novel model that incorporates an interpretable bottleneck—termed the Filter Bank (FB)—at the outset of a Variational Autoencoder (VAE). This arrangement compels the VAE to attend on the most informative segments of the input signal, fostering the learning of a novel encoding $\bold{f}_0$ which boasts enhanced interpretability and clusterability over traditional latent spaces. By deliberately constraining the VAE with this FB, we intentionally constrict its capacity to access broad input domain information, promoting the development of an encoding that is discernible, separable, and of reduced dimensionality. The evolutionary learning trajectory of $\bold{f}_0$ further manifests as a dynamic hierarchical tree, offering profound insights into cluster similarities. Additionally, for handling intricate data configurations, we propose a tailored decoder structure that is symmetrically aligned with FB's architecture. Empirical evaluations highlight the superior efficacy of ISVAE, which compares favorably to state-of-the-art results in clustering metrics across real-world datasets.
\end{abstract}



\begin{keyword}


VAE \sep time series clustering \sep interpretability \sep Human Activity Recognition

\end{keyword}

\end{frontmatter}



\section{Introduction}
Time series clustering is a technique used across various domains, such as finance \citep{miguel_23}, healthcare \citep{jamia_23}, and climate science \citep{dixit_23} to group similar time series together for its analysis and modeling. Clustering techniques offer insights into latent patterns and trends that may not be immediately apparent, which can be used for tasks such as anomaly detection, forecasting, and segmentation. Deep neural networks (DNNs) have been increasingly applied to time series clustering in recent years, with promising results. These approaches often involve encoding time series data into low-dimensional feature vectors using various embedding techniques, such as recurrent neural networks (RNNs) \citep{Sherstinsky_2020}, variational autoencoders (VAE) \citep{kingma_13}, or graph NNs (GNNs) \citep{Zhou_2018},  subsequently clustering the resulting embedding vectors using techniques such as K-means \citep{MacQueen1967SomeMF} or hierarchical clustering \citep{joeWard1963}. It is worth noting that, in most state-of-the-art approaches, learning the embedded space is typically decoupled from the clustering task itself.\\

In this paper, we introduce a novel methodology that combines unsupervised attention models with the encoding function of a VAE to construct interpretable and discriminative clustering spaces for time series. This approach is encapsulated in the proposed method, named Interpretable Spectral VAE (ISVAE).
Rather than operating on highly-correlated temporal samples, ISVAE operates on a spectral decomposition of the signal, which is obtained using the discrete-cosine transform (DCT).
Unlike most existing literature on deep clustering, ISVAE does not separate the processes of clustering and learning the embedding space. Instead, our approach simultaneously learns two interdependent feature spaces:  the traditional latent space of the VAE and an interpretable feature space. The latent space ($\mathbf{z}$) follows the conventional optimization objective of maximizing a lower bound on the signal likelihood, as the usual VAE's optimization framework. The use of this latent space as an auxiliary mechanism to optimize an interpretable feature space, inducing natural signal clustering, is one of the main contributions of ISVAE. In ISVAE, the input to the VAE's encoder is a filtered version of the input signal obtained through a bank of $L$ Gaussian spectral filters (FB), each with different central frequencies ($\mathbf{f}_0$). These frequencies are determined by an attention mechanism learned during the training of the VAE lower bound.
By means of the filter central frequencies per signal, $\mathbf{f}_0$, we establish a representation for each signal that captures the spectral bands that should be preserved to reconstruct the signal within the VAE component of the model. The feature space $\mathbf{f}_0$ can be considered a flexible clustering label, owing to the significant discretization of the encodings across different signal types (classes or clusters) in the dataset. This advantageous characteristic of ISVAE eliminates the need to determine the number of clusters in advance, which can be particularly challenging in unsupervised settings. Our experimental results illustrate the effectiveness of the proposed method in time series clustering, particularly when combining the $\mathbf{f}_0$ representation with simple clustering methods. \\

Our paper is structured as follows. Section \ref{related_work} presents an overview of state-of-the-art clustering methods. Section \ref{isvae} describes the mathematical foundation of the model, focusing on explaining the various design choices selected for each module. Section \ref{clust} characterizes the experimental setup and validation metrics. Section \ref{results} analyzes different experimental problems to provide a global picture of the ISVAE's performance, showing its key advantages through an explainability analysis based on synthetic data and comparing its performance against state-of-the-art baselines on real-world datasets.

\section{Related Work}
\label{related_work}


Clustering methods are a set of techniques used to group similar data points or objects together based on their inherent similarities, aiming to identify underlying patterns and structures in the data. Clustering algorithms are widely used in various fields and applications, such as customer segmentation \citep{wu2005research}, image analysis \citep{ng2006medical}, anomaly detection \citep{li2021clustering}, or genomic analysis \citep{hatfull2010comparative}. \\

In the context of deep learning-based clustering methods, \citep{hsu_2017} describes a novel approach for time series clustering using disentangled representations, proposing a method that involves learning a disentangled representation of the time series using a VAE, followed by a clustering algorithm to group the resulting representations. This work is extended by \citep{tonekaboni_2022}, which  generalized the previous approach, resulting in more structured latent spaces with both global and local variables. Further, \citep{ma_2019} proposes a novel unsupervised temporal representation learning model, named Deep Temporal Clustering Representation (DTCR), which integrates the temporal reconstruction and K-means objective into the seq2seq model \citep{SutskeverVL14} by reformulating it as a trace maximization problem, leading to improved cluster structures and cluster-specific temporal representations. In the context of Graph Neural Networks (GNNs) \citep{Scarselli2009TheGN} for clustering, \citep{Ferreira_2016} proposes a graph-based approach for time series clustering, which involves the construction of a time series' graph representation and the use of community detection algorithms to identify similar time series' clusters. \\

Further, various attempts to merge signal processing techniques with deep neural networks can also be found in the literature. In \citep{tamkin2020language}, a framework that uses spectral filtering for the problem of NLP was proposed, while \citep{cao2020spectral} uses the spectral domain to capture inter-series correlations and temporal dependencies jointly. In this context, \citep{moreno2023deep} proposes an attention mechanism that operates in the Fourier domain, the Spectral Attention, which merges global and local frequency domain information in the model’s embedded space. Regarding clustering, spectral clustering techniques have been largely used in applications such as computer vision \citep{yang2019deep} or source separation \citep{Trans_Neural_Networks_2006}. Unlike traditional clustering algorithms that operate in the original data space, spectral clustering leverages the spectral properties of a similarity matrix or graph Laplacian, a process that involves transforming the data into a spectral domain using techniques like the eigendecomposition of the Laplacian matrix, which captures the underlying structure and relationships within the data. This approach has proven particularly effective in handling complex or nonlinear data distributions, as it can uncover clusters that may not be well-separated in the original feature space. Moreover, the use of the Fourier transform to improve the classification and clustering performance has been largely discussed and different attempts to integrate it within more general frameworks can be found in the literature \citep{lowitz1984fourier}, \citep{goerg2011nonparametric}, \citep{holan2018time}. Nevertheless, most of the methods in the literature that incorporate signal processing techniques into deep learning architectures solely use signal decomposition techniques to identify trend and seasonality patterns, and works such as  \citep{moreno2023deep}, which operate exclusively within the frequency domain, continue to be considered as exceptions.\\

We should remark that our proposed model falls within the area of deep time-series clustering, which has seen significant research in the past decade \citep{alqahtani_21}. This line of work in clustering has not only inspired advancements in clustering itself but also in other tasks such as anomaly detection, classification, and the development of application-driven models for real-world scenarios. ISVAE draws inspiration from various architectural insights, including the utilization of auto-encoders with attention modules to enhance clustering performance \citep{olu_2021},  the analysis of frequency domain data  \citep{mu_21}, and the use of noise-reduction filters learned by deep models to improve detection in noisy communication environments \citep{z_21}. \\

In \citep{olu_2021}, an attention model with three variations is proposed, aiming to detect anomalies in electrocardiograms (ECGs). This approach employs representational learning architectures such as Auto Encoders (AE) \citep{dor_21}, Variational Autoencoders (VAE)  \citep{kingma_13}, and Long Short-Term Memory networks (LSTMs) \citep{hoch_97}. These architectures are combined with attention modules to enhance the decoder's reconstruction capability and model the distribution of outliers in ECGs through a generative process. 
In contrast to ISVAE, this approach augments the latent representation with attention scores generated by the attention module, while ISVAE applies its attention mechanism directly to the input domain.
Other methods, such as \citep{mu_21}, leverage the spectral information of the data without disregarding its temporal characteristics. They adapt both temporal and spectral attention mechanisms to focus on key frequency bands and semantically related time frames in the spectrogram.\\


Lastly, the model presented in \citep{z_21} learns a (Chebyshev) bandpass filter employing 1-D convolutional kernels (CNN), and subsequently enhancing the learned informative filters through a kernel-wise attention mechanism. In a tangentially related manner, ISVAE employs one-dimensional CNNs as feature extractors to learn the informative frequency bands selected by the attention filters, limiting the information that the VAE sees. This bottleneck, which results in interpretable feature spaces, works without supervision, contrary to the methods described in \citep{z_21}.\\

ISVAE aims to provide a comprehensive framework applicable in various scenarios, integrating all the aforementioned insights that have been utilized in a variety of models with proven performance. Our proposal offers two additional advantages: {interpretability} across the input domain and the ability to operate without the need for {supervision}. \\

\begin{figure}[b!]
    \centering
    \includegraphics[width=1\textwidth]{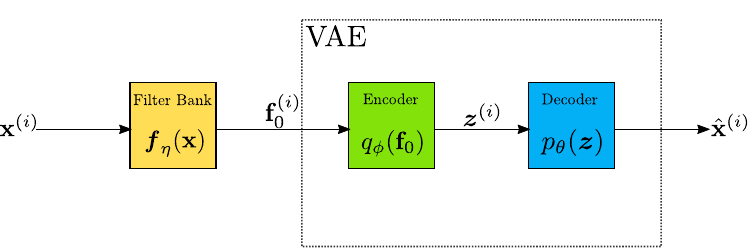}
    \caption{ISVAE Architecture.\textcolor{red}{}}
    \label{fig:arqui}
\end{figure}

\section{Interpretable Spectral VAE (ISVAE)}\label{isvae} 

As mentioned, ISVAE operates over signals in a spectral transformed space. Without loss of generality, we consider the DCT Type II transform. Let  $\mathbf{S}=\{\mathbf{s}^{(i)}\}_{i=0}^{N-1}$ be a set of $N$ real discrete-time sequences all of the same length $D$. For every signal $\mathbf{s}^{(i)}=[s_0^{(i)},\ldots,s_D^{(i)}]$, we compute its $D$-point DCT representation $\mathbf{x}^{(i)}=[x_0^{(i)},...,x_D^{(i)}]$ as follows: 
\begin{equation}
x_d^{(i)}=\sum_{n=0}^{D-1} s_n^{(i)} \cos \left[\frac{\pi}{D}\left(n+\frac{1}{2}\right) d\right] \quad \text { for } d=0, \ldots D-1.
\end{equation}
Hence, $x_d$ is the signal DCT value at a linear frequency $d/D$ relative to the sampling rate. With ISVAE, we learn to project the set $\mathbf{X} = \{\mathbf{x}^{(i)}\}^{N-1}_{i=0}$ into an alternative feature space where simple clustering methods such as K-means or DBSCAN \citep{dbscan} can be effectively applied. \\

The architecture of the ISVAE is presented in Figure \ref{fig:arqui}. Our proposal consists of two distinct stages that are interconnected. The first stage involves a Filter Bank (FB) with $J$ filters, which act as a feature extractor, reducing the input dimensionality to a vector $\mathbf{f}_0^{(i)} \in[0,1]^{J}$. The second stage employs a Variational AutoEncoder (VAE), which takes the output $\mathbf{f}_0^{(i)}$ from the FB and learns a latent space representation that is more sensitive to signal reconstruction. Unlike conventional VAE frameworks that minimize the reconstruction error between the encoder's input and the decoder's output, our VAE is trained to reconstruct the original signal $\mathbf{x}^{(i)}$ solely based on $\mathbf{f}_0^{(i)}$ as input to the encoder. Therefore, the FB is encouraged to discover a low-dimensional, high-quality representation of the original signal to facilitate the VAE's task, and attention models play a crucial role in achieving this objective. Our results demonstrate that the FB's output $\mathbf{f}_0^{(i)}$ is a suitable and informative space for clustering, surpassing the VAE's internal latent space. In the following sections, we provide a detailed description of each constituent block of the ISVAE.

\subsection{Filter Bank} \label{fb}

Consider the function\footnote{In the following, for the sake of notation, we will omit the notation $^{(i)}$.} $\mathbf{h}(f, \sigma, D) = [h_0, \ldots, h_{D-1}]$, which represents a Gaussian filter with $D$ taps. This filter has a central frequency $f$ ranging from 0 to 1, and its bandwidth is determined by $\sigma^2$. Each element $h_d$ of the filter, for $d$ ranging from 0 to $D-1$, is computed as follows:
\begin{align}
h_d = e^{-\frac{(d/D - f)^2}{2\sigma^2}}
\end{align}
When applied to an input signal $\mathbf{x}$, the filter output $\mathbf{y}$ is obtained by performing the element-wise product of $\mathbf{x}$ and $\mathbf{h}$, i.e. $\mathbf{y}=\mathbf{x}\odot\mathbf{h}$. Note that the filter has no amplitude due to simplicity, however setting the amplitude as a learnable parameter could be relevant.  \\

The Filter Bank (FB) consists of $J$ Gaussian filters $\mathbf{h}_{j} = \mathbf{h}(f_j, \sigma, D)$, where $j$ ranges from 0 to $J-1$. The hyperparameters $J$ and $\sigma$ are fixed in the ISVAE model, while the central frequencies of the filters are learned individually for each input signal $\mathbf{x}$. To determine these central frequencies, a sequential attentive architecture is proposed: $\boldsymbol{f}_{\boldsymbol{\eta}}(\mathbf{x}): \mathbb{R}_{+}^{D}\rightarrow [0,1]^{J}$. This function directly outputs the vector $\mathbf{f}_0$, being $\boldsymbol{\eta}$ the set of FB's parameters. \\

\begin{figure}[H]
    \centering
    \includegraphics[width=1\textwidth]{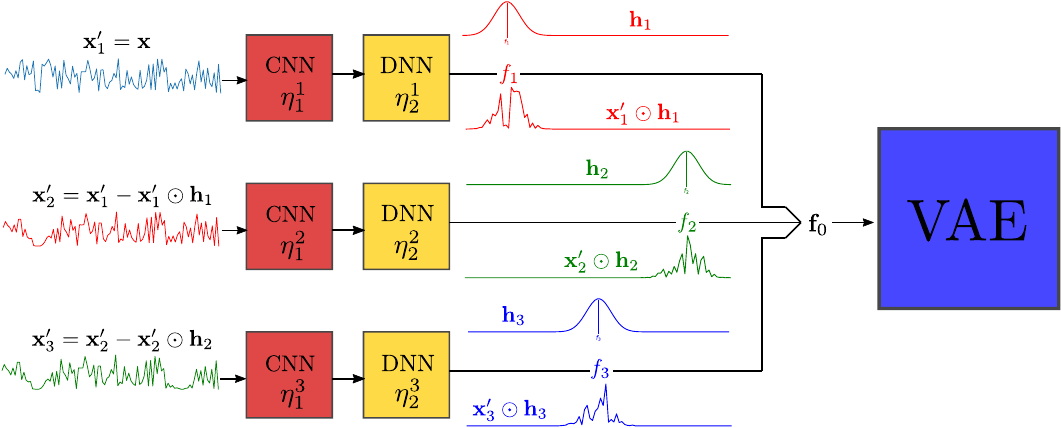}
    \caption{Sequential Filter Bank architecture }
    \label{fig:fb_arqui}
\end{figure}

 We propose a filtering scheme using the sequential procedure depicted in Figure \ref{fig:fb_arqui}. We aim to enhance the model's flexibility and promote diversity among the central frequencies of the filters. The FB consists of a series of $J$ interconnected sub-modules or branches in this sequential design. Each branch incorporates a 1D-Convolutional Neural Network (CNN) ($\mathbb{R}_+^{D}\rightarrow \mathbb{R}^{D\times L}$) with parameters $\boldsymbol{\eta}_1^j$, $j=1,\ldots,J$, to capture local correlations between adjacent frequencies. Here $L$ denotes the hidden dimension of the CNN (hyperparameter). The output of the CNN is then passed through an attentive Dense Neural Network (DNN) ($\mathbb{R}^{D\times L}\rightarrow [0,1]$) with parameters $\boldsymbol{\eta}_2^j$, $j=1,\ldots,J,$ that determines the center frequency of the $j$-th filter.  We use the same CNN and DNN structure for all branches. \\

A crucial point is how these branches are sequentially interconnected. As shown in Figure \ref{fig:fb_arqui}, the second branch does not directly observe the original signal $\mathbf{x}$ but, instead, the influence of the first filter is removed to facilitate networks in the second branch to focus on alternative regions of the spectrum. This is done at every branch, removing the influence from all the previous branches. Formally, the input at the $j$-th branch is defined as $\mathbf{x}'_j = \mathbf{x} - \sum_{k=1}^{j-1} \mathbf{h}_k \odot\mathbf{x} $, being $\mathbf{x}$ the FB's input signal.

\subsection{A VAE to reconstruct the signal given the FB output}

Once the relevant signal frequencies $\mathbf{f}_0$ have been identified for a given signal $\mathbf{x}$, we employ a VAE-style encoder-decoder module. This module aims to reconstruct $\mathbf{x}$ exclusively from $\mathbf{f}_0$. The VAE encoder distribution $q_{\boldsymbol{\boldsymbol{\phi}}}(\mathbf{z})$ ($z\in\mathbb{R}^{K}$) relies solely on $\mathbf{f}_0$, while the decoder output $p_{\boldsymbol{\theta}}(\mathbf{x}|\mathbf{z})$ defines a distribution for the complete object $\mathbf{x}$. By design, given the low-dimensional signal representation $\mathbf{f}_0$ (typically 3 to 6 dimensions in our experiments), we expect a significant reconstruction loss at the output of the VAE. However, minimizing this loss to the greatest extent possible necessitates careful selection of $\mathbf{f}_0$. Ultimately, we will utilize $\mathbf{f}_0$ for clustering purposes, making the VAE module an auxiliary component for appropriately training the FB module.\\

The encoder distribution $q_{\boldsymbol{\phi}}(\mathbf{z}|\mathbf{f}_0(\mathbf{x}))$ is assumed to be Gaussian, where a DNN with parameters $\boldsymbol{\phi}$ and input $\mathbf{f}_0$ provides the mean $\boldsymbol{\mu}(\mathbf{f}_0)$ and the diagonal covariance matrix $\boldsymbol{\Sigma}(\mathbf{f}_0)$.\\

The primary concept propelling our learning process, wherein a Variational Autoencoder (VAE) is employed as a guide to steer Filter Bank (FB) learning, results in a discrete and interpretable space $\mathbf{f}_0$. This principle is grounded in the VAE's dependency on a restricted pool of information for full signal reconstruction, thereby compelling the FB to identify and utilize informative representations. However, in practice, the intricacy of the dataset could impose an overly substantial information bottleneck on the VAE. This might hinder the VAE's ability to reconstruct the original signal effectively, leading to a static training state where further learning is obstructed. In order to address this two scenarios, we propose two decoder versions that can be interchangeable: vanilla and attentive decoder.\\

The vanilla decoder is a straightforward DNN parameterized by $\boldsymbol{\theta}$ that takes the latent code $\mathbf{z}$ and outputs the signal reconstruction $\mathbf{\hat{x}}$. Based on the results, as seen in subsequent sections, the vanilla decoder is preferable. The rationale for this preference stems from the ability of a single DNN to enhance the communication bandwidth between FB and the VAE through proximity during gradient propagation. Proximity in this context relates to the amount of layers and operations through which the gradient is back-propagated during training, thus the information passed (via gradient) from the decoder to the FB is higher when less operations/layers are between them, benefiting a more efficient cooperative learning process between both modules. Scenarios where data complexity is overtaken by the model's capacity leads to static training states where the frequency encoding $\mathbf{f}_0$ freezes along epochs.\\

For more complex scenarios, we propose the attentive decoder that is based on the idea of inducing an aid during reconstruction, creating a symmetrical architectural alignment to the FB. This is done at the cost of introducing some noise in the communication channel between FB and VAE.\\ 

\begin{figure}[H]
    \centering
    \includegraphics[width=1\textwidth]{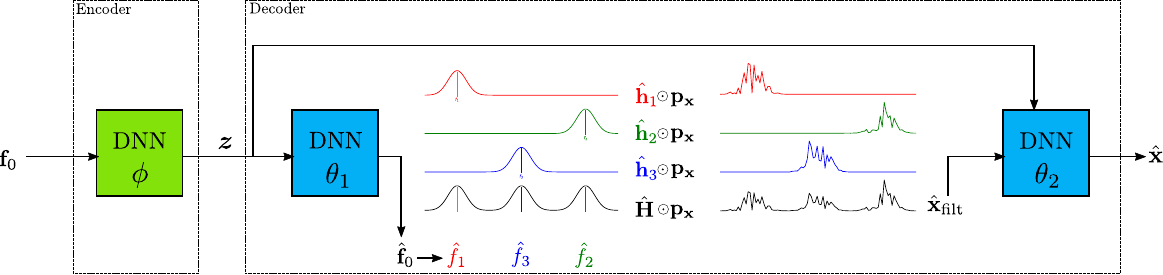}
    \caption{VAE architecture with attentive decoder}
    \label{fig:vae_arqui}
\end{figure}

 For the attentive decoder (Figure \ref{fig:vae_arqui}), we propose a nontrivial structure that employs global information about the dataset through a DCT-based averaged periodogram $\mathbf{p}_{\mathbf{X}}=\frac{1}{N} \sum_{i=1}^{N} |\mathbf{x}^{(i)}|^2$.\footnote{Although not accurate, we will refer to it as the empirical Power Spectral Density (PSD) in latter sections.} In our design, the latent sample $\mathbf{z}$ is first fed to a DNN with parameters $\boldsymbol{\theta}_1$, yielding a reconstruction of the FB's central frequencies $\hat{\mathbf{f}}_0$. The associated $J$ Gaussian filters are applied over $\mathbf{p}_{\mathbf{X}}$ and the resulting signals are combined through
\begin{align}\label{reconstruction}
    \mathbf{\hat{x}_{\text{filt}}} = \sum_{j=1}^{J}\hat{\mathbf{h}}_{j}\odot \mathbf{p}_{\mathbf{X}},
\end{align}
where $\hat{\mathbf{h}}_{j}=\mathbf{h}(\hat{f}_j, \sigma, D)$ is a Gaussian filter. $\mathbf{\hat{x}_{\text{filt}}}$ can be seen as the (magnitude-only) reconstruction of the signal $\mathbf{x}$ using the frequency bands $\hat{\mathbf{f}}_0$ (relevant only to reconstruct the signal $\mathbf{x}$) but the information on these bands is given by $\mathbf{p}_{\mathbf{X}}$.  To incorporate local information in the energy of these bands, a final DNN with parameters $\boldsymbol{\theta}_2$ combines $\mathbf{\hat{x}_{\text{filt}}}$ and $\mathbf{z}$ to construct $\boldsymbol{\mu}_{\mathbf{\hat{x}}}$, the mean of the decoder's output distribution, which again is a Gaussian pdf with fixed diagonal variance equal to $\nu$\footnote{$\nu = 1$ for all experiments in this works.}. Namely, $p_{\boldsymbol{\theta}_1,\boldsymbol{\theta}_2}(\mathbf{x}|\mathbf{z})=\mathcal{N}(\boldsymbol{\mu}_{\mathbf{\hat{x}}},\nu\mathbf{I})$. Table \ref{tab:resumen} summarizes all the networks included in ISVAE and the attentive decoder.

\begin{table}[t!]
\centering
\begin{tabular}{|lllll|}
\hline
\multicolumn{1}{|c}{\multirow{1}{*}{Filter}}&\multicolumn{1}{c}{}&\multicolumn{1}{c}{}&\multicolumn{1}{c}{CNN: $\mathbb{R}_+^{D}\rightarrow \mathbb{R}^{D\times L}$ } &\multicolumn{1}{c|}{Params: $\boldsymbol{\eta}_1^j$ }\\
\multicolumn{1}{|c}{\multirow{1}{*}{Bank}}&\multicolumn{1}{c}{$\mathbf{f}_{0}(\mathbf{x}) $}&\multicolumn{1}{c}{$[f_1,...,f_j,...,f_J]$} &\multicolumn{1}{c}{}&\multicolumn{1}{c|}{}\\
\multicolumn{1}{|c}{\multirow{1}{*}{(FB)}}&\multicolumn{1}{c}{}&\multicolumn{1}{c}{} &\multicolumn{1}{c}{DNN: $\mathbb{R}^{D\times L}\rightarrow [0,1]$}&\multicolumn{1}{c|}{Params: $\boldsymbol{\eta}^j_2$ }\\

\hline

\multirow{1}{*}{VAE} &
  \multirow{1}{*}{Encoder} &
  \multirow{1}{*}{$q_{\boldsymbol{\phi}}(\mathbf{z}|\mathbf{f}_0(\mathbf{x}))$} &
  \multirow{1}{*}{DNN: $\mathbb{R}_+^{J}\rightarrow \mathbb{R}^{K} $} & Params: $\boldsymbol{\phi}$
  \multirow{1}{*}{} \\

                     &  &  &                     &                                              \\
                       \hline

\multirow{3}{*}{VAE} &
  \multirow{3}{*}{Decoder} &
  \multirow{3}{*}{$p_{\boldsymbol{\theta}}(\mathbf{x}|\mathbf{z})$} &
  DNN: $\mathbb{R}^{K}\rightarrow \mathbb{R}^{J}$  & Params: $\boldsymbol{\theta}_1$
   \\
                     &  &  &                     &                                              \\
                     &  &  & DNN: $\mathbb{R}^{J+K}\rightarrow \mathbb{R}^{D}$ & Params: $\boldsymbol{\theta}_2$
                     \\

                     \hline
\end{tabular}
\caption{List of ISVAE neural networks and their parameters.}
\label{tab:resumen}
\end{table}

\subsection{ISVAE loss function}
\vspace{0.5cm}
The VAE evidence lower bound (ELBO) is used to optimize our model. For a data point $\mathbf{x}^{(i)}$, the ELBO is given by
\begin{equation} 
\mathcal{L}^{(i)}(\boldsymbol{\eta},\boldsymbol{\theta}, \boldsymbol{\phi})= \mathbb{E}_{q_{\boldsymbol{\phi}}(\mathbf{z}|\mathbf{f}_0(\mathbf{x}^{(i)}))}\left[\log(p_{\boldsymbol{\theta}}(\mathbf{x}^{(i)}|\mathbf{z}))\right] - D_{\mathbb{KL}}\left[q_{\boldsymbol{\phi}}(\mathbf{z}|\mathbf{f}_0(\mathbf{x}^{(i)}))||p(\mathbf{z})\right],
\label{eq:elbo}
\end{equation}
where the second term is the Kullback–Leibler divergence between the encoder distribution $q_{\boldsymbol{\phi}}(\mathbf{z}|\mathbf{f}_0(\mathbf{x}^{(i)}))$, and the prior $p(\mathbf{z})$, which is assumed to be Gaussian distributed with zero mean and identity covariance matrix. Note that the above loss functions jointly optimizes the FB parameters with the VAE encoder and decoder. Observe also that our loss function's primary purpose is to measure the reconstruction error of the full signal $\mathbf{x}$ when little information about such a signal is provided at the entry of the VAE encoder, which serves as well as a barrier to the decoder and prevents overfitting.\\


During training, the stopping criterion is to stop when the mean of the encoding $\mathbf{f}_0$ per dimension (position of each filter) does not vary in a reasonable amount of training epochs. Further, our experiments show that, for supervised datasets, this criterion is well-aligned with the clustering highest V-score during training.



\section{Clustering using ISVAE Internal Representations}\label{clust}

Clustering is not a feature inherent to VAEs, the backbone of our proposal. However, we show that the ISVAE internal representations serve as powerful signal feature spaces where clustering can effectively occur. We consider clustering using the following ISVAE's feature spaces:
\begin{enumerate}
    \item \textbf{Basic configuration}: FB central frequencies $\mathbf{f}_0(\mathbf{x}) = [f_1,\ldots,f_J]$.
    \item \textbf{Extended configuration}: FB central frequencies $\mathbf{f}_0(\mathbf{x})$ augmented with the energy of the signal $\mathbf{x}$ in each of the bands:\\
    $\left[\mathbf{f}_0(\mathbf{x}), ||\boldsymbol{h}(f_1,\sigma,D)\odot\mathbf{x}||_2^2, \ldots, \boldsymbol{h}(f_J,\sigma,D)\odot\mathbf{x}||_2^2\right]$.\footnote{The extended configuration does not require additional training since it is trivially obtained using the filter positions from the FB central frequencies.}
    \item \textbf{VAE latent codes}: samples of the VAE encoder's distribution $q_{\phi}(\mathbf{z}|\mathbf{f}_0(\mathbf{x}^{(i)}))$.\footnote{Given the superiority of the FB central frequencies, in clustering terms, latent codes are used for comparison purposes.}
\end{enumerate}

Regarding the clustering algorithms, we show remarkable results using relatively-simple methods such as K-means, DBSCAN, or spectral clustering.  To address the clustering performance, two types of metrics have been used:
 
\begin{enumerate}
    \item \textbf{Supervised metrics}: when data is labeled, we measure the consistency of the labels along the different groups using: 
    \begin{itemize}
        \item Homogeneity ($h$): measures that each cluster contains members of a single class ($h\in [0,1]$). 
        \item Completeness ($c$): measures that all members of a class are contained in the same cluster ($c\in [0,1]$).
        \item V-measure ($v$): harmonic mean between $h$ and $c$ ($v = 2\frac{h\cdot c}{h+c}$) ($v\in [0,1]$).
    \end{itemize}
    
    \item \textbf{Unsupervised metrics}: 
    \begin{itemize}
        \item Silhouette Coefficient ($s$): measures cluster's shape using the mean distance between a sample and all other points in the same class ($a$) and the former between a sample and all other data points in the next nearest cluster ($b$), according to $s = \frac{b-a}{\text{max}(a,b)}$ ($s\in [-1,1]$), being the final index the mean of each sample's metric. Higher $s$ describes better-defined clusters.
        \item Calinski-Harabsz Index ($k$): measures the ratio between the sum of inter-cluster and intra-cluster dispersion. 
        The index is higher when clusters are populated and distant from each other ($k \in [0,\infty)$).
        
    \end{itemize}
\end{enumerate}

\section{Empirical Analysis}\label{results}

In this section we provide an in-depth analysis of ISVAE's performance on different datasets. Firstly, we consider a synthetic dataset, which allows us to shed some light into ISVAE's operative. Secondly, we analyze two separate datasets that contains Human Activity Recognition (HAR) data. We first show the results obtained with the entire training set, leaving the analysis of the generalization capabilities of the model for Section \ref{validation}. 

\subsection{Synthetic Dataset}\label{synth}

We start by analyzing a synthetic dataset of $N = 1000$ signals, each of them with a dimensionality of $600$ samples.  Each signal is a mixture of $K \in \{3,4\}$  sinusoids vibrating at different frequencies (with sampling frequency $f_s = 600$) contaminated with i.i.d. Gaussian noise $\epsilon_{\text{noise}} \sim N(0.05$, $0.25$): 
\begin{equation} 
\boldsymbol{y}^{(c)} = \sum_{i=1}^K\cos\left(2\pi f_i^{(c)} \; \frac{x}{f_s}\right) + \epsilon_{\text{noise}}.
\label{eq:sinusoids}
\end{equation}
In this scenario, it is worth noting that the signals under consideration are evenly distributed among eight predefined clusters, each characterized by distinct feature vectors $f_1, \ldots, f_K$. To provide further insight into this distribution, we present the cluster-specific frequencies, as outlined in Table \ref{tab:exp1_freqs}. In this case, we implement ISVAE with only two filters, $J=2$,\footnote{We choose $J=2$ in order to visualize the $\mathbf{f}_0$ encoding space without influence of a dimensionality reduction technique such as t-SNE.} with a width of $\sigma = 15$ in both the FB and the vanilla decoder. The latent dimension is denoted $J$, i.e., $\mathbf{z} \in \mathbb{R}^2$.\\

\begin{table}[H]
    \centering
    \begin{tabular}{|l|llll|}
\cline{2-5}
\hline
 
\multicolumn{1}{|l|}{c} & $f_1$ & $f_2$ & $f_3$ & \multicolumn{1}{l|}{$f_4$ }\\ \hline
\multicolumn{1}{|l|}{$1$} & $80$  & $130$ & $495$ & -     \\ 
\multicolumn{1}{|l|}{$2$} & $180$ & $390$ & $596$ & -     \\ 
\multicolumn{1}{|l|}{$3$} & $80$  & $130$ & $230$ & $390$ \\ 
\multicolumn{1}{|l|}{$4$} & $130$ & $230$ & $430$ & $530$ \\ 
\multicolumn{1}{|l|}{$5$} & $80$  & $180$ & $315$ & $495$ \\ 
\multicolumn{1}{|l|}{$6$} & $230$ & $390$ & $495$ & $596$ \\ 
\multicolumn{1}{|l|}{$7$} & $180$ & $230$ & $430$ & $530$ \\ 
\multicolumn{1}{|l|}{$8$} & $80$  & $315$ & $495$ & $596$ \\ \hline
\end{tabular}
    \caption{Frequency components per cluster in the synthetic dataset. }
    \label{tab:exp1_freqs}
\end{table}

\begin{figure}[t!]
    \centering
    \includegraphics[width=0.7\textwidth]{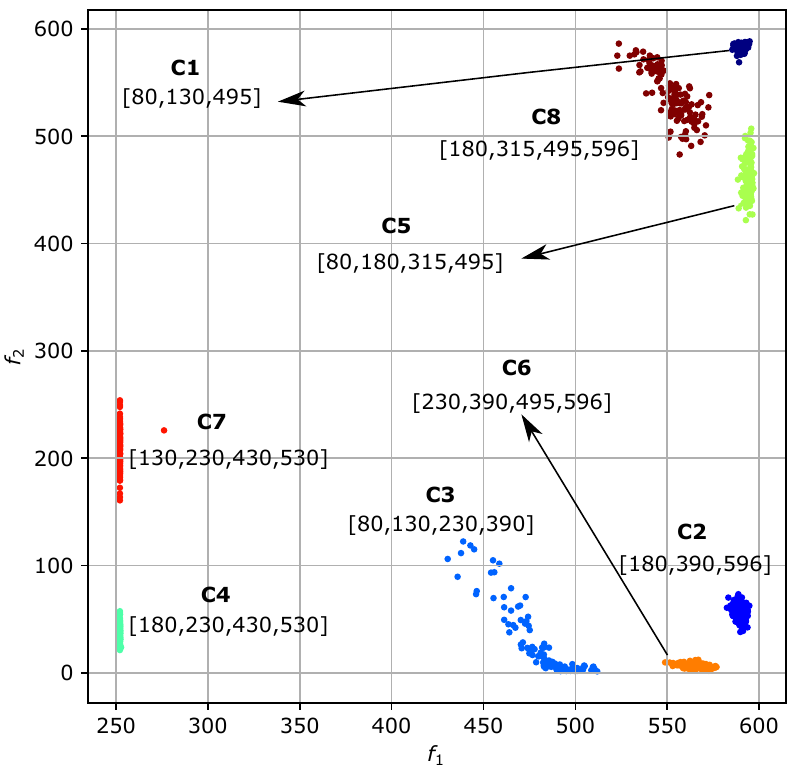}
    \caption{$\mathbf{f}_0$ space ground truth labels and cluster's generating frequencies next to each of them.}
    \label{fig:encod_syn2}
\end{figure}

When measuring similarity between two elements we must state the particular aspect from which we establish similarity. Inspecting the FB's encoding space $\mathbf{f}_0$, see Figure \ref{fig:encod_syn2}, helps understand how ISVAE is ordering the different generated signals, showing the type of similarity that is able to capture. The position of the cluster's clouds in the encoding space is more organized and discrete compared to that found in the latent space, showed and discussed in Annex B (\ref{annexc}). Each sample encoding $\mathbf{f}_0 = [f_1, f_2]$ can be viewed as a point in the plane, and the clouds of points seems to be arranged based on proximity regarding the shared frequencies between clusters. One might assume that the encoding should capture the most representative and exclusive frequencies with respect to other clusters or render a unique combination of those, exclusive to each class. However, the pointed frequencies discovered by the FB do not align with the generating frequencies of the dataset.The model instead creates separate encoding regions, compressing the spectral information in a new representation, which is not completely separated from the frequency positions manifested at the PSD. In these regions, the frequency information for each group is included in proximity to its related clusters, while also taking into account the spectral information of the entire dataset.

\begin{figure}[t!]
    \centering
    \includegraphics[width=1\textwidth]{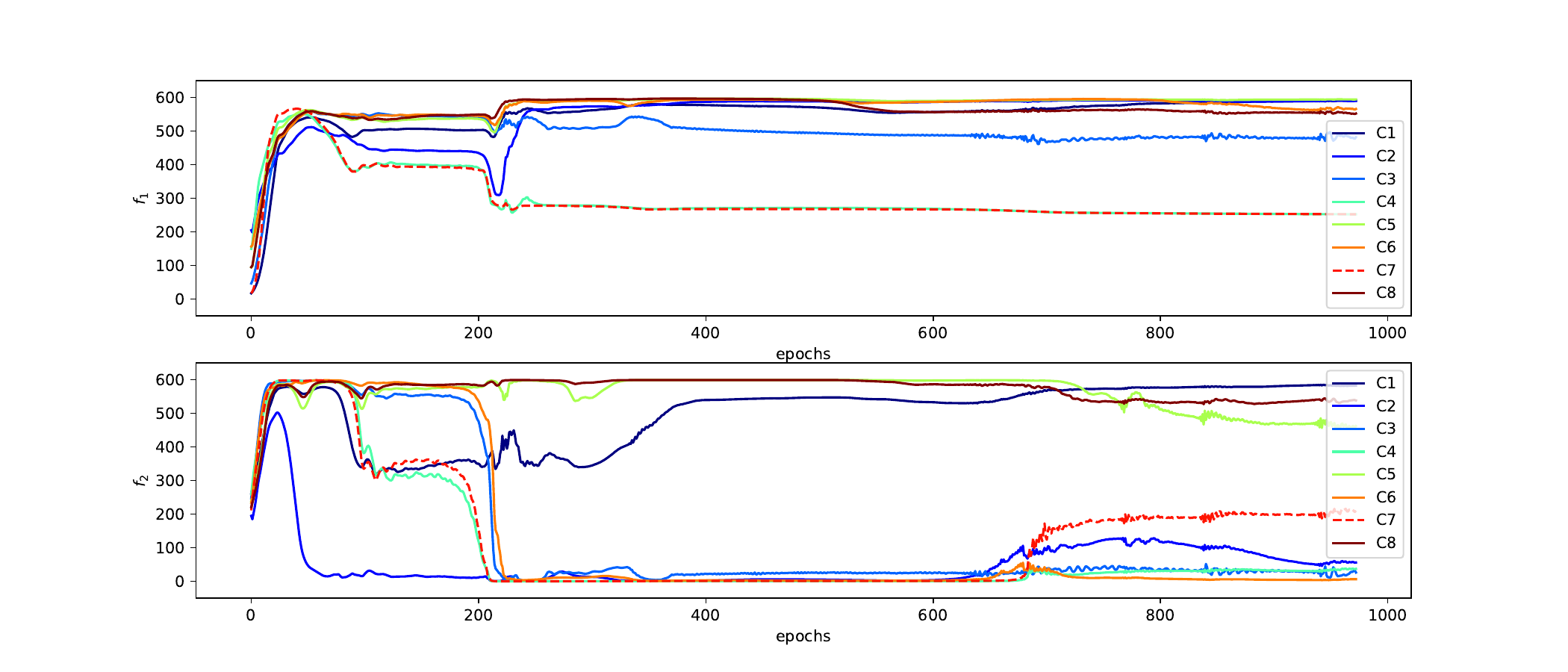}
    \caption{Evolution of filters per dimension and per cluster during training. Each line represents the mean central frequency across each class.}
    \label{fig:evo_syn2}
\end{figure}

Two key observations must be considered jointly to comprehend these findings. The first insight is that the encoding process is not directed by a preconceived objective accurately tuned to find discretized solutions. Instead, it is entirely a consequence of minimizing the VAE's Evidence Lower Bound (ELBO), where reconstruction quality and non-overfitted solutions are optimized. Secondly and looking at several training realizations, the multi-modality observed at a certain encoding dimension is notably higher than the other one. We argue that each of the different encoding dimensions learned by ISVAE  prioritize learning of local vs global features independently and under a trade-off scenario between such perspectives of the dataset. In the current example, $f_2$ is prioritizing local aspects of the signals while $f_1$ is considering more general characteristics, since the multi-modality manifested in the latest is higher and less concentrated than the previous (Figure \ref{fig:evo_syn2}). A good example supporting the previous claims are clusters $4$ and $7$, where $f_1$ reserves a common position for both given the high number of shared frequencies ($[230,430,530]$) while $f_2$ separates the local part of each of them. However, it is not always possible to render such a clear separation between global and local parts thus reaching an equilibrium point in the trade-off problem. Such behavior is aligned with the architectural philosophy of the FB where upper/shallower branches omit characteristics for the deeper/lower filters. 

Figure \ref{fig:evo_syn2} illustrates the (mean) progression of filter allocation per class throughout training. Such process can be seen as dynamic hierarchical tree, with each branch successively partitioning into less dense cluster groupings until an independent path forms for each cluster. Considering this, the ISVAE model seems to capture spectral similarity between signals by condensing several elements such as global versus local feature distinction, spectral proximity of resonating frequencies, and the ease of reconstructing the original signal from a limited filter set, effectively serving as a bottleneck.\\

The interpretability trait that gives name to our model emerges from the combination of indirect and direct outcomes that ISVAE can produce. This feature facilitates a blended understanding of local and global characteristics of the dataset under study, enhancing overall analysis efficacy.

\subsection{Human Activity Recognition (HAR)} \label{har}
\vspace{0.5cm}

In this section, we evaluate the performance of ISVAE on real-world data. Specifically, we use two datasets derived from the Human Activity Recognition (HAR) dataset presented in \citep{Leut_13}. This HAR dataset includes data from seven individuals, captured using two three-axis accelerometers positioned on their waist and ankle, recorded simultaneously. We utilized just the first axis data from the waist accelerometer. Despite this restriction, the exclusion of the other axes does not significantly impact the results, as demonstrated by the success of the clustering task.\\

The subjects were engaged in five distinct activities --sitting, lying, standing, walking, and running-- carried out under semi-naturalistic conditions for a minimum duration of 20 minutes. These activities were manually annotated by referencing video footage, and the original recordings were down-sampled to 16 Hz. We curated two datasets for our analysis: one dataset including activity transitions (fully-unsupervised) and another excluding them (supervised). It is important to note that the term ``supervised'' here refers solely to the dataset creation process, not to the information that ISVAE had access to, as in both cases, ISVAE was not privy to supervised information. The generation process of the two datasets was to divide recordings in $7$ second windows ( $\mathbf{x} \in \mathbb{R}^{112}$), with no distinction among subjects. For the supervised dataset, windows where a unique activity was present were taken into consideration, and for the unsupervised case, all windows were considered.\\


%



In the following sections, we will apply ISVAE to both datasets and assess its performance. Vanilla and attentive decoders will be used for the supervised and unsupervised HAR dataset respectively, although in Annex's \ref{annexb} and \ref{annexc} a comparison of both variations will be addressed for each experimental setting. No remarkable differences in performance are seen in the current HAR datasets.  Furthermore, we will also evaluate the extended configuration, which notably requires no additional training time. 

\subsubsection{Supervised HAR dataset (without transitions)}
\vspace{0.5cm}

In terms of hyperparameter selection, we selected a three-dimensional latent space ($\mathbf{z} \in \mathbb{R}^3$) and consider $J = 6$ filters, each with a bandwidth of $\sigma = 6$. We use the same selection for both decoder choices. The model was trained over a total of $300$ epochs. $20$ realizations where executed, concurrently ran using both basic and extended clustering configurations. As previously stated, the extended clustering configuration utilizes a concatenation of additional deterministic features, built on the basic configuration ($\mathbf{f}_0$). This strategy avoids incurring in additional training time and ensures a fair comparison between the clustering configurations.\\

For each realization, the chosen metric for comparison was the highest Variance Score (V-score) obtained during training.\footnote{Note that the highest V-scores of the basic and extended configuration is obtained at different points during training.}

\begin{figure}[H]
\begin{adjustbox}{width=1.5\columnwidth,center}
\begin{tabular}{ccc}

\includegraphics[width=0.15\textwidth]{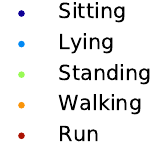}&\includegraphics[width=0.425\textwidth]{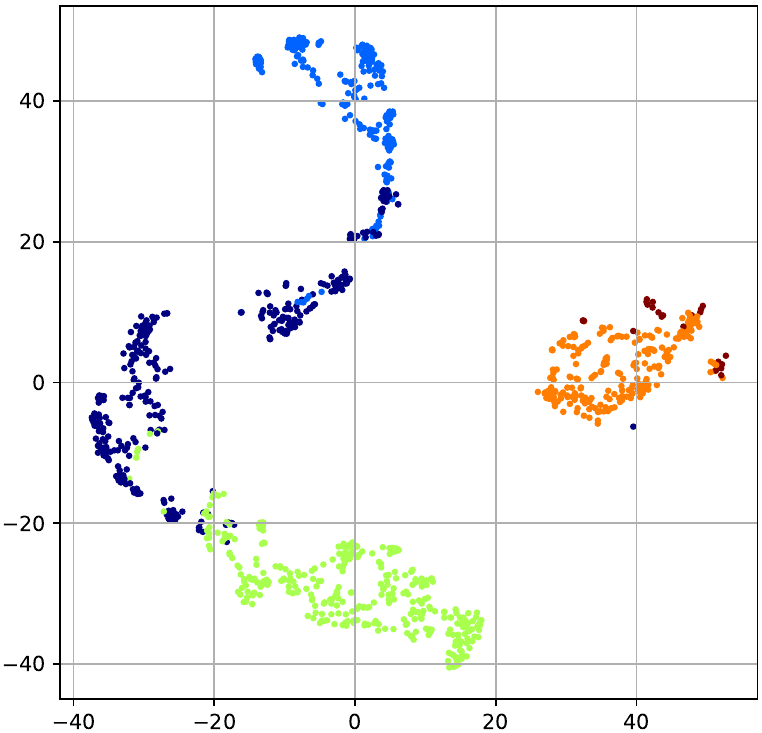}&\includegraphics[width=0.425\textwidth]{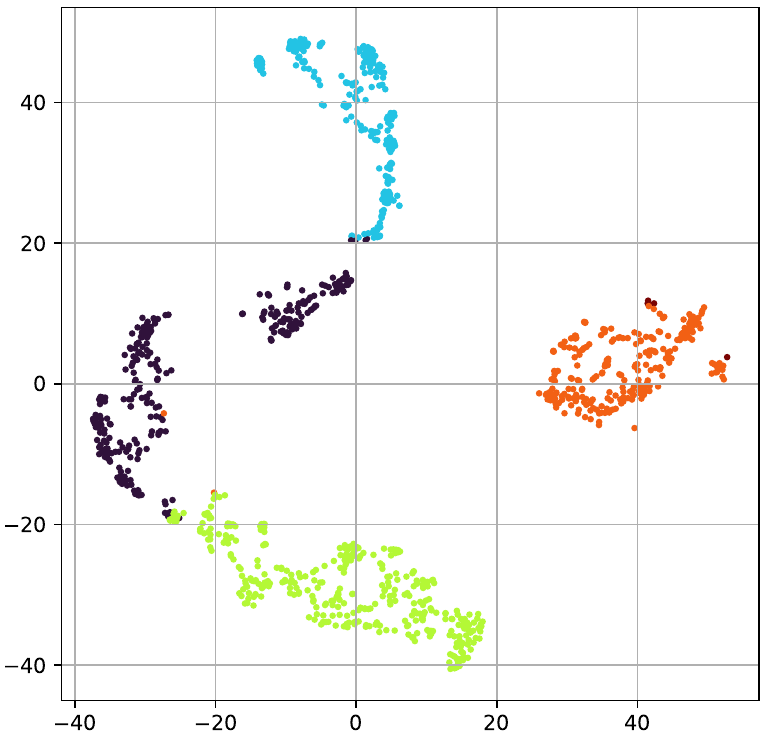}
\end{tabular}
\end{adjustbox}

\caption{\textbf{t-SNE} projection of $\mathbf{f}_0$ (ext) using the \textbf{vanilla} decoder. Left: ground truth labels. Right: K-Means predicted labels (V-score $= 0.8138$)}
\label{fig:simple_dec_har_main}
\end{figure}

Given the slightly improved performance (mean over the 20 realizations) of the vanilla decoder against the attentive decoder and avoiding selection bias, in Figure \ref{fig:simple_dec_har_main} we show the t-SNE projection of the learned  $\mathbf{f}_0$ encoding obtained at realization $3$. Left plot uses color coding associated to ground truth labels and the right one shows predicted labels obtained using the extended configuration $\mathbf{f}_0$ (ext).\footnote{An important remark is that t-SNE projection is only computed using the central frequencies, leaving the associated energy from each filter (extended configuration) solely for label prediction on the clustering task.}\\

 Quantitatively, the V-score (0.8138) points out to an outstanding clusterization, taking into account ISVAE has only access to a single gyroscope's axis even with imbalanced classes like `running' (shown in brown in the ground truth labels). Secondly, and qualitatively, the relative location of classes seems to follow an understandable pattern. Given that the depicted points are t-SNE projections which preserve the proximity of points in the original space to some extent, the clusters are organized according to the rate of movement, effectively captured by the frequency information of the signals. For example, the `running' and `walking' clusters are always in proximity, regardless of the realization.\\

This dataset does not differentiate between subjects and, at the individual level, the unique behaviors within a single class can generate variability. However, ISVAE seems to capture the global characteristics of the signals, enabling a successful distinction between activities. Moreover, within a single activity, the model is also learning differences, evidenced by the positional discontinuity of points within a single class. An example of this is the `sitting' activity, shown in dark blue color in Figure \ref{fig:simple_dec_har_main}, where three roughly compact clusters are separated without affecting class detection. Additionally, it should be noted that ISVAE, even without supervision, is capable of separating all classes, as seen in the ground truth. This occurs even if K-means doesn't detect them accurately.\\

\subsubsection{Unsupervised HAR dataset \textbf{with} transitions}
\vspace{0.5cm}
In this section a more realistic scenario to prove ISVAE robustness is analyzed, where various activities (including transitions) can be present in a single window ($\mathbf{x} \in \mathbb{R}^{112}$). In general terms, this scenario is a harder battleground for any clustering method due to its highly imbalanced proportion among labels, specially transitions; and also the unsupervised condition, which hinders the ability to model groups with a unique structure as opposed to the prior case without transitions, where groups held less variable members in absence of any other activity being present in the same time window. Given the fact that human error is present in the manual process of labeling activities, where the determination of the beginning and end of the transitions to other activities is not accurate enough, taking an unsupervised approach surprisingly yields a successful clusterization of the main activities. In this way, the current dataset can be seen as noisy version of the previous one, holding up to 10 classes. For visualization and supervised metric calculation the most repeated (mode) activity is considered as ground truth.\\

\begin{figure}[H]
\begin{adjustbox}{width=1.5\columnwidth,center}
\begin{tabular}{ccc}
\includegraphics[width=0.17\textwidth]{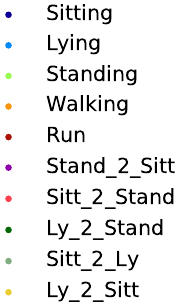}&\includegraphics[width=0.425\textwidth]{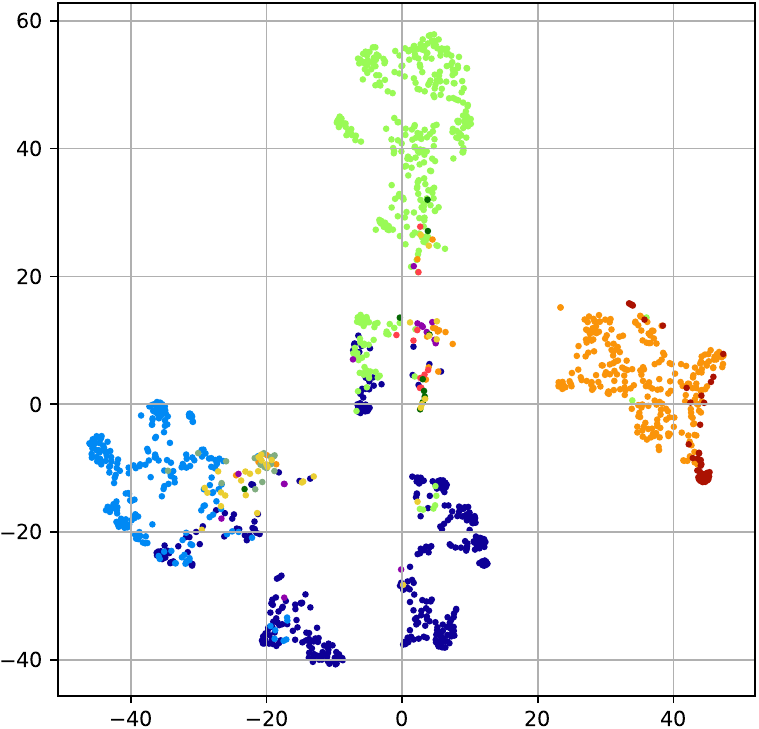}&\includegraphics[width=0.425\textwidth]{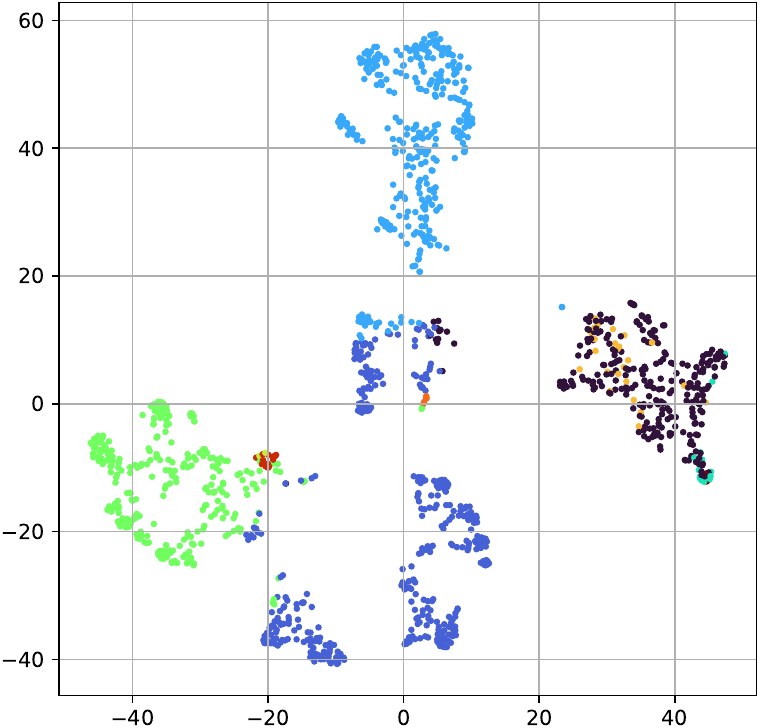}
\end{tabular}
\end{adjustbox}

\caption{\textbf{t-SNE} projection of $\mathbf{f}_0$ (ext) using the \textbf{attentive} decoder. Left: ground truth labels. Right: K-Means predicted labels (V-Score $=0.6842$)}
\label{fig:complex_dec_har_trans_main}
\end{figure}

Focusing on the qualitative aspect, Figure \ref{fig:complex_dec_har_trans_main} shows the t-SNE projection of encoding $\mathbf{f}_0$ of realization $1$ using the attentive decoder. Again the predicted clustering labels are obtained using the extended configuration. The result manifests remarkable performance not only separating main classes but also transitions in an interpretable way. Encoding space discretely separates the main activities in different regions at the same time that captures intra-class variability as seen in the supervised dataset, in spite of the noisy information. Looking at the ground truth labels, transitions between activities are still not confidently distinguished by K-means in the encoding space but are rather concentrated in a unique region or placed in between activities that define their transition, hinting that ISVAE understands transitions as outliers with common characteristics of the neighboring clouds. A good example is the bidirectional transition between lying and sitting  ('Ly\_2\_Sitt' and 'Sitt\_2\_Ly' labels), in which ISVAE generally locates them in between the borders of both activities.\\

Naturally, given the high amount of imbalanced classes and the shortened duration of signal transitions, K-means ran over the encoding space discovers new groups within a single activity (walking activity) and classifies different transitions into a single cluster, yielding a substantially lower score ($0.6842$) if compared to the supervised case.

\subsubsection{Performance comparison using different hyperparameters and several datasets\\}\label{validation}

The primary aim of this section is to provide a comparative analysis between various baselines and all model variations (across all its variations), utilizing both a validation and a test set, in order to obtain a comprehensive understanding of the model's performance. Our attention has been concentrated on the Human Activity Recognition Task, incorporating a total of five datasets that vary by the type of sensors used and the nature of the classes, thereby encompassing a broad array of activities. Table \ref{tab:data_sum} provides a summary of each dataset's characteristics. Notably, ISVAE outperforms all baselines across all datasets, suggesting a distinct advantage conferred by the decoder architecture, which varies depending on the specific dataset. The NN arquitecture details for all experiments (including the synthetic one) can be found in Appendix \ref{annexf}
\begin{table}[b!]
\centering
\begin{tabular}{|c|cccc|}
\hline
\multicolumn{1}{|c|}{Dataset} &
  Dimensionality &
  $N$ &
  Number of activities &
\multicolumn{1}{l|}{Reference} \\
  \hline

HAR &
  $\mathbf{x} \in \mathbb{R}^{112}$ &
  1108 &
  5 &
  \citep{Leut_13} \\
HAR (transitions) &
  $\mathbf{x} \in \mathbb{R}^{112}$ &
  1344 &
  10 &
  \citep{Leut_13} \\
Active HAR  &
  $\mathbf{x} \in \mathbb{R}^{125}$ &
  600 &
  5&
  \citep{HAR_act} \\
SoDA (accelerometer) &
  $\mathbf{x} \in \mathbb{R}^{100}$ &
  549 &
  6 &
  \citep{wang2022social} \\
SoDA (gyroscope) &
  $\mathbf{x} \in \mathbb{R}^{100}$ &
  549 &
  6 &
 \citep{wang2022social}\\
 \hline
\end{tabular}
\caption{Summary of datasets}
\label{tab:data_sum}
\end{table}

Each of the aforementioned datasets has undergone a unique preprocessing procedure. The generation pipeline for the first two datasets (HAR with and without transitions) has been described in detail in Section \ref{har}.\\

Next, we created a new dataset from \citep{HAR_act}'s, which we term as `HAR Active', reflecting the inclusion of high-frequency movement of 5 classes: standing, walking, running, jumping, and cycling. Our objective is to evaluate ISVAE's ability to distinguish between various sports activities. The original dataset contains a total of 19 daily and sports activities, recorded from 8 different subjects using five different body placements (torso, right and left arms, and legs). Each body part recorded signals simultaneously using an accelerometer, gyroscope, and magnetometer, all with 3-axis (x,y, and z) resolution. Similarly to the previous HAR datasets, our generated dataset does not differentiate between subjects and only incorporates the y-axis of the magnetometer from the left leg, considering ISVAE's handling of single-dimensionality. We selected the magnetometer to address ISVAE's interaction with a wide variety of sensors, while the choice of the y-axis was arbitrary. The dataset maintains a perfectly balanced representation of classes, with each class comprising 120 time signals (with a posterior frequency transformation), each lasting 5 seconds, at a sampling rate of 25 Hz (resulting in 125 data points per signal).\\

Finally and utilizing the SoDA dataset from ``Social Distancing Alert with Smartwatches'' \citep{wang2022social}, two datasets have been generated, each recorded using an accelerometer or a gyroscope, with the goal of comparing sensor performance under ISVAE over a single activity set. The original dataset contains 19 complex activities that where chosen in the context of social distancing under the COVID-19 virus pandemic, from which we filter six varied classes (hugging, high five, walking, running, drink water, key stroking).  The complexity of the dataset allows us to characterize ISVAE's clustering performance when fed to quasi-periodic (walking and running) and non periodic activity patterns (rest). The original dataset contains 10 repetitions of each activity performed by 10 volunteers. In order to homogenize signal duration, adapting it to ISVAE requirements, a fixed time window is chosen, resulting in 100 data points per signal. \\

The experimental setting is consistent across all datasets. Datasets have been divided into train (75\%), test (12.5\%), and validation (12.5\%) partitions, each appropriately normalized using a standard scaler. The metrics are computed as the mean and standard deviation of 6 realizations over 1000 epochs\footnote{We choose a sufficiently large training period to ensure the best effort on cluster separability by the model.}, applied to all clustering configurations and ISVAE's decoder architectures. The model at the epoch with the maximum V-score obtained in validation is the one used to evaluate the different clustering metrics over the test partition. ISVAE is tested using three different values for the number of filters ($J \in \{4,5,6\}$), with a fixed filter bandwidth for each dataset that is adapted according to its input dimensionality. K-means is applied to either the learned latent space or the encoding space ($\textbf{f}_0$), in both basic and extended configurations. A simple VAE was also tested using the same criteria, with K-means applied to its learned latent space.\\

In cases where K-means was applied directly to the original signals, in both time and frequency domains, the number of realizations was increased to 100, considering the relatively low computational resource demand for these cases.

\begin{table}[H]
\begin{adjustbox}{width=1\columnwidth,center}
\begin{tabular}{cc|ccccc|}
\cline{3-7}
\multicolumn{1}{l}{\multirow{2}{*}{}} &
  \multicolumn{1}{l}{\multirow{2}{*}{}} &
  \multicolumn{5}{|c|}{\multirow{2}{*}{Dataset \#1: HAR (no transitions)}} \\
\multicolumn{1}{l}{} &
  \multicolumn{1}{l|}{} &
  \multicolumn{5}{c|}{} \\ \cline{3-7} 
\multicolumn{1}{l}{} &
  \multicolumn{1}{l|}{} &
  \multicolumn{5}{c|}{K-Means} \\ \cline{2-7} 
\multicolumn{1}{l|}{} &
  \multicolumn{1}{l|}{J} &
  V-score &
  Homogeneity  &
  Completeness  &
  Silhouette &
  \multicolumn{1}{c|}{Calinski-Harabasz} \\ \hline
\multicolumn{1}{|c|}{Time} &
  \multicolumn{1}{c|}{-} &
  $0.6870\pm0.0037$ &
  $0.6165\pm0.0047$ &
  $0.7758\pm0.0099$ &
  $0.3625\pm0.0483$ &
  $180.89\pm9.66$ \\ \hline
\multicolumn{1}{|c|}{DCT} &
  \multicolumn{1}{c|}{-} &
  $0.0143\pm0.0590$ &
  $0.0088\pm0.0406$ &
  $0.1727\pm0.2741$ &
  $0.2437\pm0.3957$ &
  $11.25\pm20.96$ \\ \hline
\multicolumn{1}{|c|}{Vanilla VAE ($z$)} &
  \multicolumn{1}{c|}{-} &
  $0.6027\pm0.1380$ &
  $0.5572\pm0.1559$ &
  $0.6774\pm0.1193$ &
  $0.6514\pm0.0988$ &
  $1401.11\pm796.03$ \\ \hline
\multicolumn{1}{|c|}{\multirow{3}{*}{\begin{tabular}{c}ISVAE\\ vanila dec.\\  ($z$)\end{tabular}}} &
  $4$ &
  $0.6571\pm0.0549$ &
  $0.6763\pm0.0614$ &
  $0.6394\pm0.0504$ &
  $0.4994\pm0.0710$ &
  $1334.88\pm437.62$ \\
\multicolumn{1}{|c|}{} &
   $5$ &
  $0.7140\pm0.0807$ &
  $0.7098\pm0.0866$ &
  $0.7258\pm0.1085$ &
  $0.6317\pm0.0726$ &
  $1504.71\pm908.48$ \\
\multicolumn{1}{|c|}{} &
  $6$ &
  $0.6319\pm0.2779$ &
  $0.6394\pm0.2831$ &
  $0.6266\pm0.2761$ &
  $0.5820\pm0.1733$ &
  $1002.77\pm696.01$ \\ \hline
\multicolumn{1}{|c|}{\multirow{3}{*}{\begin{tabular}[c]{@{}c@{}}ISVAE\\ vanila dec.\\ $\mathbf{f}_0$\end{tabular}}} &
   &
  $0.7234\pm0.1381$ &
  $0.6965\pm0.1815$ &
  $0.7702\pm0.0724$ &
  $0.6643\pm0.0206$ &
  $226.06\pm81.94$ \\
\multicolumn{1}{|c|}{} &
   &
  $0.7087\pm0.0641$ &
  $0.65.12\pm0.0386$ &
  $0.7803\pm0.0988$ &
  $0.6143\pm0.0892$ &
  $174.69\pm101.29$ \\
\multicolumn{1}{|c|}{} &
   &
  $\bold{0.8325\pm0.0841}$ &
  $\bold{0.8096\pm0.1269}$ &
  $0.8633\pm0.0351$ &
  $0.6750\pm0.0567$ &
  $283.15\pm161.38$ \\ \hline
\multicolumn{1}{|c|}{\multirow{3}{*}{\begin{tabular}[c]{@{}c@{}}ISVAE\\ vanila dec.\\ $\mathbf{f}_0$ (ext)\end{tabular}}} &
   &
  $0.7127\pm0.1648$ &
  $0.6246\pm0.2080$ &
  $0.8728\pm0.0631$ &
  $0.6403\pm0.0478$ &
  $135.55\pm38.74$ \\
\multicolumn{1}{|c|}{} &
   &
  $0.7151\pm0.0863$ &
  $0.7102\pm0.0863$ &
  $0.7274\pm0.1068$ &
  $0.6330\pm0.0702$ &
  $\bold{1505.75\pm907.70}$ \\
\multicolumn{1}{|c|}{} &
   &
  $0.8306\pm0.0729$ &
  $0.7600\pm0.1148$ &
  $\bold{0.9269\pm0.0523}$ &
  $0.6218\pm0.0403$ &
  $103.41\pm26.61$ \\ \hline
\multicolumn{1}{|c|}{\multirow{3}{*}{\begin{tabular}[c]{@{}c@{}}ISVAE\\ attentive dec.\\ ($z$)\end{tabular}}} &
   &
  $0.5522\pm0.1246$ &
  $0.5674\pm0.1274$ &
  $0.5382\pm0.1229$ &
  $0.4696\pm0.1324$ &
  $786.11\pm1182.33$ \\
\multicolumn{1}{|c|}{} &
   &
  $0.6504\pm0.1547$ &
  $0.6523\pm0.1540$ &
  $0.6503\pm0.1596$ &
  $0.5815\pm0.0742$ &
  $693.21\pm346.91$ \\
\multicolumn{1}{|c|}{} &
   &
  $0.7175\pm0.1156$ &
  $0.7172\pm0.1218$ &
  $0.7183\pm0.1108$ &
  $0.5445\pm0.1234$ &
  $435.01\pm297.64$ \\ \hline
\multicolumn{1}{|c|}{\multirow{3}{*}{\begin{tabular}[c]{@{}c@{}}ISVAE\\ attentive dec.\\ $\mathbf{f}_0$\end{tabular}}} &
   &
  $0.7777\pm0.0610$ &
  $0.7556\pm0.1032$ &
  $0.8096\pm0.0337$ &
  $\bold{0.6819\pm0.0259}$ &
  $249.12\pm125.97$ \\
\multicolumn{1}{|c|}{} &
   &
  $0.7216\pm0.1934$ &
  $0.6769\pm0.2190$ &
  $0.7821\pm0.1708$ &
  $0.6734\pm0.0834$ &
  $174.77\pm78.63$ \\
\multicolumn{1}{|c|}{} &
   &
  $0.7442\pm0.0499$ &
  $0.6837\pm0.0795$ &
  $0.8220\pm0.0338$ &
  $0.6216\pm0.0296$ &
  $103.05\pm28.91$ \\ \hline
\multicolumn{1}{|c|}{\multirow{3}{*}{\begin{tabular}[c]{@{}c@{}}ISVAE\\ attentive dec.\\ $\mathbf{f}_0$ (ext)\end{tabular}}} &
   &
  $0.7463\pm0.0974$ &
  $0.6610\pm0.1393$ &
  $0.8867\pm0.0378$ &
  $0.6246\pm0.0432$ &
  $118.49\pm22.08$ \\
\multicolumn{1}{|c|}{} &
   &
  $0.7188\pm0.1187$ &
  $0.6123\pm0.1359$ &
  $0.8880\pm0.0543$ &
  $0.6262\pm0.0230$ &
  $100.87\pm18.40$ \\
\multicolumn{1}{|c|}{} &
   &
  $0.7526\pm0.0699$ &
  $0.6509\pm0.0623$ &
  $0.8930\pm0.0844$ &
  $0.5645\pm0.0621$ &
  $84.61\pm13.54$ \\ \hline
\end{tabular}
\end{adjustbox}
\caption{Test results for HAR dataset without transitions using K-means}
\label{tab:exp_HAR}
\end{table}

\begin{table}[H]
\begin{adjustbox}{width=1\columnwidth,center}
\begin{tabular}{|cc|ccccc|}
\cline{3-7}
\multicolumn{1}{l}{\multirow{2}{*}{}} &
  \multicolumn{1}{l|}{\multirow{2}{*}{}} &
  \multicolumn{5}{c|}{\multirow{2}{*}{Dataset \#2: HAR (transitions)}} \\
\multicolumn{1}{l}{} &
  \multicolumn{1}{l|}{} &
  \multicolumn{5}{c|}{} \\ \cline{3-7} 
\multicolumn{1}{l}{} &
  \multicolumn{1}{l|}{} &
  \multicolumn{5}{c|}{K-Means}   \\ \cline{2-7}
\multicolumn{1}{l|}{} &
  \multicolumn{1}{l|}{J} &
  V-score &
  Homogeneity  &
  Completeness  &
  Silhouette &
  Calinski-Harabasz \\ \hline
\multicolumn{1}{|l|}{Time} &
  \multicolumn{1}{c|}{-} &
  $0.6115\pm0.0111$ &
  $0.5498\pm0.0239$ &
  $0.6909\pm0.0244$ &
  $0.3285\pm0.0596$ &
  $140.1497\pm17.9285$ \\ \hline
\multicolumn{1}{|l|}{DCT} &
  \multicolumn{1}{c|}{-} &
  $0.0283\pm0.0825$ &
  $0.0174\pm0.0552$ &
  $0.3143\pm0.2416$ &
  $0.5350\pm0.3967$ &
  $18.3147\pm20.6660$ \\ \hline
\multicolumn{1}{|l|}{Vanilla VAE ($z$)} &
  \multicolumn{1}{c|}{-} &
  $0.4304\pm 0.1391$ &
  $0.4409\pm0.1732$ &
  $0.4276\pm0.1118$ &
  $0.4408\pm0.0576$ &
  $840.83\pm409.67$ \\ \hline
\multicolumn{1}{|c|}{\multirow{3}{*}{\begin{tabular}[c]{@{}c@{}}ISVAE\\ vanila dec.\\  ($z$)\end{tabular}}} &
   $4$ &
  $0.5927\pm0.0647$ &
  $0.6601\pm0.0619$ &
  $0.5383\pm0.0669$ &
  $0.5139\pm0.1057$ &
  $\bold{2154.6310\pm2060.8637}$ \\
\multicolumn{1}{|c|}{} &
   $5$ &
  $0.6160\pm0.0577$ &
  $0.6908\pm0.0626$ &
  $0.5559\pm0.0539$ &
  $0.4836\pm0.0611$ &
  $1994.1955\pm743.7120$ \\
\multicolumn{1}{|c|}{} &
   $6$ &
  $0.6552\pm0.0204$ &
  $0.6994\pm0.0439$ &
  $0.6185\pm0.0273$ &
  $0.5657\pm0.0660$ &
  $1423.5126\pm623.9857$ \\ \hline
\multicolumn{1}{|c|}{\multirow{3}{*}{\begin{tabular}[c]{@{}c@{}}ISVAE\\ vanila dec.\\ $\mathbf{f}_0$\end{tabular}}} &
   &
  $0.7028\pm0.0910$ &
  $0.7411\pm0.0943$ &
  $0.6690\pm0.0909$ &
  $0.5725\pm0.0730$ &
  $393.5489\pm238.4036$ \\
\multicolumn{1}{|c|}{} &
   &
  $0.7482\pm0.0470$ &
  $0.7856\pm0.0736$ &
  $0.7166\pm0.0333$ &
  $0.5941\pm0.0635$ &
  $244.3843\pm72.7358$ \\
\multicolumn{1}{|c|}{} &
   &
  $\bold{0.7635\pm0.0538}$ &
  $\bold{0.7861\pm0.0851}$ &
  $0.7454\pm0.0283$ &
  $0.5974\pm0.075$ &
  $161.2769\pm49.5783$ \\ \hline
\multicolumn{1}{|c|}{\multirow{3}{*}{\begin{tabular}[c]{@{}c@{}}ISVAE\\ vanila dec.\\ $\mathbf{f}_0$ (ext)\end{tabular}}} &
   &
  $0.7277\pm0.1003$ &
  $0.7205\pm0.1271$ &
  $0.7404\pm0.0836$ &
  $0.5622\pm0.0882$ &
  $228.0998\pm72.8796$ \\
\multicolumn{1}{|c|}{} &
   &
  $0.6191\pm0.0577$ &
  $0.6955\pm0.0588$ &
  $0.5580\pm0.0563$ &
  $0.4896\pm0.0575$ &
  $2019.0210\pm754.9356$ \\
\multicolumn{1}{|c|}{} &
   &
  $0.7455\pm0.0660$ &
  $0.7098\pm0.0982$ &
  $\bold{0.7903\pm0.0314}$ &
  $0.5637\pm0.0441$ &
  $114.8484\pm30.4362$ \\ \hline
\multicolumn{1}{|c|}{\multirow{3}{*}{\begin{tabular}[c]{@{}c@{}}ISVAE\\ attentive dec.\\ ($z$)\end{tabular}}} &
   &
  $0.5066\pm0.0622$ &
  $0.5769\pm0.0670$ &
  $0.4517\pm0.0580$ &
  $0.4007\pm0.0729$ &
  $299.9807\pm200.7653$ \\
\multicolumn{1}{|c|}{} &
   &
  $0.6420\pm0.0378$ &
  $0.6964\pm0.0215$ &
  $0.5972\pm0.0549$ &
  $0.5420\pm0.0917$ &
  $930.1067\pm1081.4253$ \\
\multicolumn{1}{|c|}{} &
   &
  $0.6129\pm0.0611$ &
  $0.6747\pm0.0524$ &
  $0.5620\pm0.0673$ &
  $0.4738\pm0.0685$ &
  $569.6354\pm368.0223$ \\ \hline
\multicolumn{1}{|c|}{\multirow{3}{*}{\begin{tabular}[c]{@{}c@{}}ISVAE\\ attentive dec.\\ $\mathbf{f}_0$\end{tabular}}} &
   &
  $0.7349\pm0.0410$ &
  $0.7516\pm0.0557$ &
  $0.7199\pm0.0327$ &
  $\bold{0.6544\pm0.0240}$ &
  $253.2825\pm49.6228$ \\
\multicolumn{1}{|c|}{} &
   &
  $0.7140\pm0.0430$ &
  $0.7217\pm0.0489$ &
  $0.7070\pm0.0412$ &
  $0.5971\pm0.0549$ &
  $207.5511\pm84.0253$ \\
\multicolumn{1}{|c|}{} &
   &
  $0.7217\pm0.0447$ &
  $0.7404\pm0.0707$ &
  $0.7061\pm0.0283$ &
  $0.5458\pm0.0574$ &
  $134.6388\pm62.8179$ \\ \hline
\multicolumn{1}{|c|}{\multirow{3}{*}{\begin{tabular}[c]{@{}c@{}}ISVAE\\ attentive dec.\\ $\mathbf{f}_0$ (ext)\end{tabular}}} &
   &
  $0.7200\pm0.0886$ &
  $0.6877\pm0.1154$ &
  $0.7609\pm0.0596$ &
  $0.6405\pm0.0177$ &
  $180.8839\pm53.1447$ \\
\multicolumn{1}{|c|}{} &
   &
  $0.6964\pm0.0629$ &
  $0.6543\pm0.0922$ &
  $0.7515\pm0.0374$ &
  $0.5925\pm0.0387$ &
  $152.7447\pm48.6932$ \\
\multicolumn{1}{|c|}{} &
   &
  $0.6911\pm0.0621$ &
  $0.6542\pm0.0997$ &
  $0.7428\pm0.0355$ &
  $0.5710\pm0.0291$ &
  $99.8168\pm25.8532$ \\ \hline
\end{tabular}
\end{adjustbox}
\caption{Test results for HAR dataset with transitions using K-means}
\label{tab:exp_HAR_trans}
\end{table}

\begin{table}[H]
\begin{adjustbox}{width=1\columnwidth,center}
\begin{tabular}{|cc|ccccc|}
\cline{3-7}
\multicolumn{1}{l}{\multirow{2}{*}{}} &
  \multicolumn{1}{l|}{\multirow{2}{*}{}} &
  \multicolumn{5}{c|}{\multirow{2}{*}{Dataset \#3: Active HAR}} \\
\multicolumn{1}{l}{} &
  \multicolumn{1}{l|}{} &
  \multicolumn{5}{c|}{} \\ \cline{2-7} 
\multicolumn{1}{l|}{} &
  \multicolumn{1}{l|}{} &
  \multicolumn{5}{c|}{K-Means} \\ \cline{2-7} 
\multicolumn{1}{l|}{} &
  \multicolumn{1}{l|}{J} &
  \multicolumn{1}{c}{V-score} &
  \multicolumn{1}{c}{Homogeneity} &
  \multicolumn{1}{c}{Completeness} &
  \multicolumn{1}{c}{Silhouette} &
  \multicolumn{1}{c|}{Calinski-Harabasz} \\ \hline
\multicolumn{1}{|c|}{Time} &
  \multicolumn{1}{c|}{-} &
  $0.5683\pm0.0337$ &
  $0.4705\pm0.0241$ &
  $0.7213\pm0.0710$ &
  $0.2560\pm0.0166$ &
  $12.85\pm0.64$ \\ \hline
\multicolumn{1}{|c|}{DCT} &
  \multicolumn{1}{c|}{-} &
  $0.1608\pm0.2063$ &
  $0.1068\pm0.1424$ &
  $0.3722\pm0.3918$ &
  $0.1209\pm0.1420$ &
  $12.67\pm17.06$ \\ \hline
\multicolumn{1}{|c|}{Vanilla VAE ($z$)} &
  \multicolumn{1}{c|}{-} &
  $0.3027\pm0.0217$ &
  $0.2973\pm0.0212$ &
  $0.3084\pm0.0224$ &
  $0.3112\pm0.0100$ &
  $35.07\pm2.39$ \\ \hline
\multicolumn{1}{|c|}{\multirow{3}{*}{\begin{tabular}[c]{@{}c@{}}ISVAE\\ vanila dec.\\  ($z$)\end{tabular}}} &
  $4$ &
  $0.2980\pm0.0753$ &
  $0.2937\pm0.0733$ &
  $0.3024\pm0.0775$ &
  $0.3112\pm0.0282$ &
  $37.20\pm5.48$ \\
\multicolumn{1}{|c|}{} &
  $5$ &
  $0.3073\pm0.0929$ &
  $0.3028\pm0.0910$ &
  $0.3121\pm0.0951$ &
  $0.2939\pm0.0269$ &
  $38.12\pm5.20$ \\ 
\multicolumn{1}{|c|}{} &
  $6$ &
  $0.3342\pm0.0377$ &
  $0.3284\pm0.0342$ &
  $0.3404\pm0.0421$ &
  $0.3001\pm0.0137$ &
  $35.38\pm3.88$ \\ \hline
\multicolumn{1}{|c|}{\multirow{3}{*}{\begin{tabular}[c]{@{}c@{}}ISVAE\\ vanila dec.\\ $\mathbf{f}_0$\end{tabular}}} &
   &
  $0.6754\pm0.0976$ &
  $0.6505\pm0.1038$ &
  $0.7042\pm0.0919$ &
  $0.5494\pm0.0559$ &
  $200.41\pm131.73$ \\
\multicolumn{1}{|c|}{} &
   &
  $0.7156\pm0.0484$ &
  $0.6879\pm0.0471$ &
  $0.7460\pm0.0527$ &
  $0.5009\pm0.0585$ &
  $93.06\pm31.85$ \\
\multicolumn{1}{|c|}{} &
   &
  $0.6990\pm0.0562$ &
  $0.6650\pm0.0463$ &
  $0.7373\pm0.0708$ &
  $0.5522\pm0.0642$ &
  $44.39\pm111.15$ \\ \hline
\multicolumn{1}{|c|}{\multirow{3}{*}{\begin{tabular}[c]{@{}c@{}}ISVAE\\ vanila dec.\\ $\mathbf{f}_0$ (ext)\end{tabular}}} &
   &
  $0.7937\pm0.1241$ &
  $0.7568\pm0.1432$ &
  $0.8375\pm0.1038$ &
  $0.4900\pm0.0694$ &
  $72.7953\pm35.3560$ \\
\multicolumn{1}{|c|}{} &
   &
  $\bold{0.8526\pm0.0965}$ &
  $\bold{0.8182\pm0.1201}$ &
  $\bold{0.8938\pm0.0763}$ &
  $0.4993\pm0.0506$ &
  $62.8914\pm15.3287$ \\
\multicolumn{1}{|c|}{} &
   &
  $0.8144\pm0.0731$ &
  $0.7897\pm0.0795$ &
  $0.8414;\pm0.0698$ &
  $0.4518\pm0.0559$ &
  $52.7908\pm19.6117$ \\ \hline
\multicolumn{1}{|c|}{\multirow{3}{*}{\begin{tabular}[c]{@{}c@{}}ISVAE\\ attentive dec.\\ ($z$)\end{tabular}}} &
   &
  $0.2855\pm0.0714$ &
  $0.2801\pm0.0702$ &
  $0.2912\pm0.0730$ &
  $0.3043\pm0.0253$ &
  $38.03\pm7.04$ \\
\multicolumn{1}{|c|}{} &
   &
  $0.2060\pm0.1076$ &
  $0.2020\pm0.1057$ &
  $0.2103\pm0.1095$ &
  $0.2868\pm0.0237$ &
  $33.17\pm3.96$ \\
\multicolumn{1}{|c|}{} &
   &
  $0.3017\pm0.1051$ &
  $0.2979\pm0.1036$ &
  $0.3057\pm0.1068$ &
  $0.2936\pm0.0116$ &
  $38.78\pm5.03$ \\ \hline
\multicolumn{1}{|c|}{\multirow{3}{*}{\begin{tabular}[c]{@{}c@{}}ISVAE\\ attentive dec.\\ $\mathbf{f}_0$\end{tabular}}} &
   &
  $0.7314\pm0.0689$ &
  $0.7203\pm0.0643$ &
  $0.7428\pm0.0743$ &
  $\bold{0.5575\pm0.0611}$ &
  $\bold{194.81\pm130.91}$ \\
\multicolumn{1}{|c|}{} &
   &
  $0.6404\pm0.0790$ &
  $0.6191\pm0.1026$ &
  $0.6680\pm0.0471$ &
  $0.5144\pm0.0925$ &
  $120.69\pm 83.41$ \\
\multicolumn{1}{|c|}{} &
   &
  $0.6788\pm0.0638$ &
  $0.6553\pm 0.0685$ &
  $0.7045\pm0.0594$ &
  $0.4999\pm0.0581$ &
  $87.04\pm 53.64$ \\ \hline
\multicolumn{1}{|c|}{\multirow{3}{*}{\begin{tabular}[c]{@{}c@{}}ISVAE\\ attentive dec.\\ $\mathbf{f}_0$ (ext)\end{tabular}}} &
   &
  $0.8376\pm0.0764$ &
  $0.8111\pm0.0915$ &
  $0.8676\pm0.0595$ &
  $0.4855\pm0.0570$ &
  $66.09\pm20.69$ \\
\multicolumn{1}{|c|}{} &
   &
  $0.7191\pm0.0412$ &
  $0.6800\pm0.0432$ &
  $0.7644\pm0.0489$ &
  $0.4114\pm0.0431$ &
  $44.11\pm15.92$ \\
\multicolumn{1}{|c|}{} &
   &
  $0.7746;\pm0.0998$ &
  $0.7405\pm0.1162$ &
  $0.8135;\pm0.0811$ &
  $0.4352\pm0.0431$ &
  $40.58\pm11.18$ \\ \hline
\end{tabular}
\end{adjustbox}
\caption{Test results for HAR active dataset using K-means}
\label{tab:exp_HAR_act}

\end{table}


\begin{table}[H]
\begin{adjustbox}{width=1\columnwidth,center}
\begin{tabular}{|cc|ccccc|}
\cline{3-7}
\multicolumn{1}{l}{\multirow{2}{*}{}} &
  \multicolumn{1}{l|}{\multirow{2}{*}{}} &
  \multicolumn{5}{c|}{\multirow{2}{*}{Dataset \#4: SoDA (accelerometer)}} \\
\multicolumn{1}{l}{} &
  \multicolumn{1}{l|}{} &
  \multicolumn{5}{c|}{} \\ \cline{2-7} 
\multicolumn{1}{l|}{} &
  \multicolumn{1}{l|}{} &
  \multicolumn{5}{c|}{K-Means} \\ \cline{2-7} 
\multicolumn{1}{l|}{} &
  \multicolumn{1}{l|}{J} &
  \multicolumn{1}{c}{V-score} &
  \multicolumn{1}{c}{Homogeneity} &
  \multicolumn{1}{c}{Completeness} &
  \multicolumn{1}{c}{Silhouette} &
  \multicolumn{1}{c|}{Calinski-Harabasz} \\ \hline
\multicolumn{1}{|c|}{Time} &
  \multicolumn{1}{c|}{-} &
  $0.7081\pm0.0206$ &
  $0.6629\pm0.0189$ &
  $0.7602\pm0.0265$ &
  $0.4104\pm0.0076$ &
  $54.66\pm2.06$ \\ \hline
\multicolumn{1}{|c|}{DCT} &
  \multicolumn{1}{c|}{-} &
  $0.0768\pm0.1342$ &
  $0.0505\pm0.0898$ &
  $0.1956\pm0.2670$ &
  $0.1136\pm0.2095$ &
  $1.79\pm2.72$ \\ \hline
\multicolumn{1}{|c|}{Vanilla VAE ($z$)} &
  \multicolumn{1}{c|}{-} &
  $0.7163\pm0.0539$ &
  $0.6632\pm0.0595$ &
  $0.7794\pm0.0478$ &
  $0.5967\pm0.0585$ &
  $266.24\pm178.23$ \\ \hline
\multicolumn{1}{|c|}{\multirow{3}{*}{\begin{tabular}[c]{@{}c@{}}ISVAE\\ vanila dec.\\  ($z$)\end{tabular}}} &
  $4$ &
  $0.6912\pm0.0334$ &
  $0.6406\pm0.0350$ &
  $0.7506\pm0.0319$ &
  $0.6562\pm0.0296$ &
  $688.26\pm454.18$ \\
\multicolumn{1}{|c|}{} &
  $5$ &
  $0.5976\pm0.2379$ &
  $0.5530\pm0.2206$ &
  $0.6505\pm0.2591$ &
  $0.5762\pm0.1419$ &
  $359.25\pm154.79$ \\
\multicolumn{1}{|c|}{} &
  $6$ &
  $0.6121\pm0.1033$ &
  $0.5594\pm0.1046$ &
  $0.6764\pm0.0997$ &
  $\bold{0.6681\pm0.0363}$ &
  $\bold{659.26\pm236.75}$ \\ \hline
\multicolumn{1}{|c|}{\multirow{3}{*}{\begin{tabular}[c]{@{}c@{}}ISVAE\\ vanila dec.\\ $\mathbf{f}_0$\end{tabular}}} &
   &
  $0.6632\pm0.1259$ &
  $0.6097\pm0.1273$ &
  $0.7281\pm0.1217$ &
  $0.5595\pm0.0896$ &
  $105.29\pm63.71$ \\
\multicolumn{1}{|c|}{} &
   &
  $0.6913\pm0.0953$ &
  $0.6310\pm0.1010$ &
  $0.7677\pm0.0940$ &
  $0.5329\pm0.0693$ &
  $60.86\pm20.22$ \\
\multicolumn{1}{|c|}{} &
   &
  $0.6681\pm0.1021$ &
  $0.6185\pm0.0993$ &
  $0.7272\pm0.1075$ &
  $0.4991\pm0.0898$ &
  $51.21\pm15.02$ \\ \hline
\multicolumn{1}{|c|}{\multirow{3}{*}{\begin{tabular}[c]{@{}c@{}}ISVAE\\ vanila dec.\\ $\mathbf{f}_0$ (ext)\end{tabular}}} &
   &
  $0.5380\pm0.0791$ &
  $0.4494\pm0.0801$ &
  $0.6743\pm0.0749$ &
  $0.4733\pm0.0863$ &
  $36.05\pm9.72$ \\
\multicolumn{1}{|c|}{} &
   &
  $0.6381\pm0.1228$ &
  $0.5662\pm0.1240$ &
  $0.7339\pm0.1191$ &
  $0.4622\pm0.0827$ &
  $28.21\pm8.14$ \\
\multicolumn{1}{|c|}{} &
   &
  $0.5818\pm0.0763$ &
  $0.5156\pm0.0835$ &
  $0.6704\pm0.0668$ &
  $0.4488\pm0.0607$ &
  $27.34\pm3.60$ \\ \hline
\multicolumn{1}{|c|}{\multirow{3}{*}{\begin{tabular}[c]{@{}c@{}}ISVAE\\ attentive dec.\\ ($z$)\end{tabular}}} &
   &
  $0.6491\pm0.1074$ &
  $0.6062\pm0.1007$ &
  $0.6988\pm0.1162$ &
  $0.5686\pm0.1056$ &
  $399.00\pm46.79$ \\
\multicolumn{1}{|c|}{} &
   &
  $0.5332\pm0.1168$ &
  $0.4956\pm0.1153$ &
  $0.5774\pm0.1172$ &
  $0.5445\pm0.0966$ &
  $231.14\pm191.23$ \\
\multicolumn{1}{|c|}{} &
   &
  $0.5967\pm0.1157$ &
  $0.5504\pm0.1142$ &
  $0.6525\pm0.1175$ &
  $0.5616\pm0.0703$ &
  $221.79\pm91.23$ \\ \hline
\multicolumn{1}{|c|}{\multirow{3}{*}{\begin{tabular}[c]{@{}c@{}}ISVAE\\ attentive dec.\\ $\mathbf{f}_0$\end{tabular}}} &
   &
  $0.7225\pm0.0797$ &
  $0.6656\pm0.0871$ &
  $0.7912\pm0.0692$ &
  $0.6151\pm0.0919$ &
  $140.67\pm73.38$ \\
\multicolumn{1}{|c|}{} &
   &
  $0.6285\pm0.0785$ &
  $0.5809\pm0.0772$ &
  $0.6850\pm0.0805$ &
  $0.5051\pm0.0377$ &
  $65.82\pm18.50$ \\
\multicolumn{1}{|c|}{} &
   &
  $\bold{0.7450\pm0.0852}$ &
  $\bold{0.6855\pm0.0886}$ &
  $\bold{0.8163\pm0.0797}$ &
  $0.5062\pm0.0346$ &
  $54.67\pm10.49$ \\ \hline
\multicolumn{1}{|c|}{\multirow{3}{*}{\begin{tabular}[c]{@{}c@{}}ISVAE\\ attentive dec.\\ $\mathbf{f}_0$ (ext)\end{tabular}}} &
   &
  $0.5860\pm0.1053$ &
  $0.5153\pm0.1030$ &
  $0.6804\pm0.1052$ &
  $0.5064\pm0.0460$ &
  $45.36\pm9.33$ \\
\multicolumn{1}{|c|}{} &
   &
  $0.6863\pm0.1319$ &
  $0.6153\pm0.1370$ &
  $0.7794\pm0.1200$ &
  $0.4600\pm0.0594$ &
  $33.30\pm4.95$ \\
\multicolumn{1}{|c|}{} &
   &
  $0.5711\pm0.0648$ &
  $0.5036\pm0.0753$ &
  $0.6634\pm0.0547$ &
  $0.4329\pm0.0473$ &
  $143.63\pm30.28$ \\ \hline
\end{tabular}
\end{adjustbox}
\caption{Test results for SODA (accelerometer) dataset using K-means}
\label{tab:exp_SODA_acc}
\end{table}

\begin{table}[H]
\begin{adjustbox}{width=1\columnwidth,center}
\begin{tabular}{|cc|ccccc|}
\cline{3-7}
\multicolumn{1}{l}{\multirow{2}{*}{}} &
  \multicolumn{1}{l|}{\multirow{2}{*}{}} &
  \multicolumn{5}{c|}{\multirow{2}{*}{Dataset \#4: SoDA (gyroscope)}} \\
\multicolumn{1}{l}{} &
  \multicolumn{1}{l|}{} &
  \multicolumn{5}{c|}{} \\ \cline{2-7} 
\multicolumn{1}{l|}{} &
  \multicolumn{1}{l|}{} &
  \multicolumn{5}{c|}{K-Means} \\ \cline{2-7} 
\multicolumn{1}{l|}{} &
  \multicolumn{1}{l|}{J} &
  \multicolumn{1}{c}{V-score} &
  \multicolumn{1}{c}{Homogeneity} &
  \multicolumn{1}{c}{Completeness} &
  \multicolumn{1}{c}{Silhouette} &
  \multicolumn{1}{c|}{Calinski-Harabasz} \\ \hline
\multicolumn{1}{|c|}{Time} &
  \multicolumn{1}{c|}{-} &
  $0.2459\pm0.0437$ &
  $0.1818\pm0.0399$ &
  \textbf{$0.3872\pm0.0359$} &
  $0.0637\pm0.1222$ &
  $7.15\pm0.65$ \\  \hline
\multicolumn{1}{|c|}{DCT} &
  \multicolumn{1}{c|}{-} &
  $0.0256\pm0.0366$ &
  $0.0156\pm0.0235$ &
  $0.1212\pm0.1262$ &
  $0.1451\pm0.2088$ &
  $1.84\pm2.18$ \\  \hline
\multicolumn{1}{|c|}{Vanilla VAE ($z$)} &
  \multicolumn{1}{c|}{-} &
  $0.2600\pm0.0536$ &
  $0.2091\pm0.0494$ &
  $\bold{0.3462\pm0.0599}$ &
  $0.4263\pm0.0696$ &
  $44.41\pm6.58$ \\  \hline
\multicolumn{1}{|c|}{\multirow{3}{*}{\begin{tabular}[c]{@{}c@{}}ISVAE\\ vanila dec.\\  ($z$)\end{tabular}}} &
  $4$ &
  $0.2238\pm0.0704$ &
  $0.1953\pm0.0584$ &
  $0.2627\pm0.0885$ &
  $0.6035\pm0.0437$ &
  $350.59\pm174.64$ \\
\multicolumn{1}{|c|}{} &
  $5$ &
  $0.1942\pm0.0702$ &
  $0.1759\pm0.0647$ &
  $0.2170\pm0.0768$ &
  $\bold{0.5728\pm0.0642}$ &
  $\bold{355.99\pm235.52}$ \\
\multicolumn{1}{|c|}{} &
  $6$ &
  $0.1815\pm0.0801$ &
  $0.1566\pm0.0693$ &
  $0.2167\pm0.0947$ &
  $0.5660\pm0.1121$ &
  $257.53\pm93.62$ \\ \hline
\multicolumn{1}{|c|}{\multirow{3}{*}{\begin{tabular}[c]{@{}c@{}}ISVAE\\ vanila dec.\\ $\mathbf{f}_0$\end{tabular}}} &
   &
  $0.2206\pm0.0698$ &
  $0.2022\pm0.0653$ &
  $0.2433\pm0.0766$ &
  $0.5562\pm0.0760$ &
  $59.37\pm15.06$ \\
\multicolumn{1}{|c|}{} &
   &
  $0.2211\pm0.0727$ &
  $0.1921\pm0.0728$ &
  $0.2648\pm0.0739$ &
  $0.4392\pm0.0739$ &
  $37.54\pm10.65$ \\
\multicolumn{1}{|c|}{} &
   &
  $0.1578\pm0.0330$ &
  $0.1377\pm0.0299$ &
  $0.1861\pm0.0405$ &
  $0.3211\pm0.0490$ &
  $23.07\pm4.81$ \\ \hline
\multicolumn{1}{|c|}{\multirow{3}{*}{\begin{tabular}[c]{@{}c@{}}ISVAE\\ vanila dec.\\ $\mathbf{f}_0$ (ext)\end{tabular}}} &
   &
  $0.2267\pm0.0325$ &
  $0.1892\pm0.0284$ &
  $0.2871\pm0.0536$ &
  $0.3231\pm0.0516$ &
  $18.77\pm2.42$ \\
\multicolumn{1}{|c|}{} &
   &
  $0.2564\pm0.0903$ &
  $0.2102\pm0.0803$ &
  $0.3379\pm0.1146$ &
  $0.3016\pm0.0781$ &
  $15.58\pm2.77$ \\
\multicolumn{1}{|c|}{} &
   &
  $0.2522\pm0.0743$ &
  $0.2059\pm0.0653$ &
  $0.3276\pm0.0861$ &
  $0.2055\pm0.0308$ &
  $11.30\pm1.25$ \\ \hline
\multicolumn{1}{|c|}{\multirow{3}{*}{\begin{tabular}[c]{@{}c@{}}ISVAE\\ attentive dec.\\ ($z$)\end{tabular}}} &
   &
  $0.1421\pm0.0341$ &
  $0.1337\pm0.0321$ &
  $0.1517\pm0.0363$ &
  $0.2776\pm0.0153$ &
  $27.89\pm2.77$ \\
\multicolumn{1}{|c|}{} &
   &
  $0.1128\pm0.0464$ &
  $0.1053\pm0.0440$ &
  $0.1215\pm0.0491$ &
  $0.2775\pm0.0187$ &
  $29.95\pm4.62$ \\
\multicolumn{1}{|c|}{} &
   &
  $0.1378\pm0.0359$ &
  $0.1245\pm0.0298$ &
  $0.1546\pm0.0448$ &
  $0.2971\pm0.0406$ &
  $33.27\pm7.08$ \\ \hline
\multicolumn{1}{|c|}{\multirow{3}{*}{\begin{tabular}[c]{@{}c@{}}ISVAE\\ attentive dec.\\ $\mathbf{f}_0$\end{tabular}}} &
   &
  $0.1993\pm0.0830$ &
  $0.1796\pm0.0747$ &
  $0.2245\pm0.0940$ &
  $0.5674\pm0.1344$ &
  $54.73\pm20.00$ \\
\multicolumn{1}{|c|}{} &
   &
  $0.1751\pm0.0872$ &
  $0.1478\pm0.0731$ &
  $0.2155\pm0.1086$ &
  $0.5031\pm0.0753$ &
  $38.26\pm5.44$ \\
\multicolumn{1}{|c|}{} &
   &
  $0.1649\pm0.0585$ &
  $0.1413\pm0.0503$ &
  $0.2000\pm0.0725$ &
  $0.2957\pm0.0606$ &
  $21.47\pm4.92$ \\ \hline
\multicolumn{1}{|c|}{\multirow{3}{*}{\begin{tabular}[c]{@{}c@{}}ISVAE\\ attentive dec.\\ $\mathbf{f}_0$ (ext)\end{tabular}}} &
   &
  $0.2183\pm0.0655$ &
  $0.1882\pm0.0627$ &
  $0.2638\pm0.0709$ &
  $0.3141\pm0.0771$ &
  $18.61\pm2.19$ \\
\multicolumn{1}{|c|}{} &
   &
  $0.2274\pm0.0961$ &
  $0.1882\pm0.0859$ &
  $0.2942\pm0.1078$ &
  $0.3069\pm0.0834$ &
  $15.85\pm2.37$ \\
\multicolumn{1}{|c|}{} &
   &
  $\bold{0.2670\pm0.0995}$ &
  $\bold{0.2205\pm0.0836}$ &
  $0.3407\pm0.1253$ &
  $0.2111\pm0.0612$ &
  $11.37\pm1.24$ \\ \hline
\end{tabular}
\end{adjustbox}
\caption{Test results for SODA (gyroscope) dataset using K-means}
\label{tab:exp_SODA_gyr}
\end{table}

In general terms, ISVAE's generalizes to unseen data, not portraying any overfitting problems. Such behavior is expected since, by design, overfitting with our model is a difficult task since signal at reconstruction is remarkably noisy. The infused bottleneck is responsible to limit information at the decoder making the reconstruction error high if compared to a vanilla VAE. Most importantly, results show that the encoding learned by ISVAE is remarkably improved for clustering that the latent space obtained in a vanilla VAE.  \\

Regarding the results in both supervised and unsupervised HAR datasets (Tables \ref{tab:exp_HAR} and \ref{tab:exp_HAR_trans}), the best V-score  is achieved with the vanilla decoder using $6$ filters and performing clustering over the basic $\boldsymbol{f}_0$ encoding. Similar to other datasets, non supervised clustering metrics (silhouette and Calinski-Harabasz scores) are obtained with different configurations, making it difficult to address ISVAE's configuration optimality in unsupervised settings where ground truth labels are not available. Nevertheless, high V-scores are obtained with such ISVAE configurations. \\

ISVAE's performance in the Active HAR dataset (Table \ref{tab:exp_HAR_act}) converges at its maximum with the extended configuration $\boldsymbol{f}_0$ (ext.) using the vanilla decoder, achieving a V-score of $0.8526$ over the test partition.\\

Finally, attentive decoder is optimal for both SODA datasets (accelerometer \ref{tab:exp_SODA_acc} and gyroscope \ref{tab:exp_SODA_gyr}) given the dataset complexity. On one hand using the accelerometer data, a V-score of $0.7450$ is obtained when clustering over the basic configuration encoding $\boldsymbol{f}_0$. On the other hand with gyroscope data, a small V-score of $0.2670$ slightly overpasses clustering with a vanilla VAE. Results over the SODA datasets manifest the usefulness of the attentive decoder when dealing with highly complex data with low separability between groups. \\

In order to address performance using other clustering methods, we repeated the above procedure with 500 epochs per realization and compute the mean V-score over 5 realizations over the test set. Optimal configurations may not coincide with the above experiments due to the reduction in training time by a factor of $2$. For DBSCAN we use an $\epsilon=0.5$ in all datasets. In Table \ref{tab:clust_HAR_exp}, V-score is showed for K-Means, DBSCAN and Spectral clustering using the HAR (no transitions). The maximum score is always obtained with K-Means for all datasets, suggesting ISVAE's encoding space $\bold{f}_0$ is spherical in nature. In Annex \ref{annexg}, tables for the rest of datasets are shown.

\begin{table}[b!]
\begin{adjustbox}{width=1\columnwidth,center}
\begin{tabular}{cc|ccc|}
\cline{3-5}
\multicolumn{1}{l}{\multirow{2}{*}{}} &
  \multicolumn{1}{l|}{\multirow{2}{*}{}} &
  \multicolumn{3}{c|}{\multirow{2}{*}{Dataset \#1: HAR (no transitions)}} \\
\multicolumn{1}{l}{} &
  \multicolumn{1}{l|}{} &
  \multicolumn{3}{c|}{} \\ \cline{2-5} 
\multicolumn{1}{l|}{} &
  \multicolumn{1}{l|}{} &
  \multicolumn{3}{c|}{V-score} \\ \cline{2-5} 
\multicolumn{1}{l|}{} &
  \multicolumn{1}{l|}{J} &
  \multicolumn{1}{c|}{K-Means} &
  \multicolumn{1}{c|}{DBSCAN} &
  \multicolumn{1}{c|}{Spectral} \\ \hline
\multicolumn{1}{|l|}{Time} &
  \multicolumn{1}{c|}{-} &
  $0.6856\pm 0.0126$ &
  $\bold{0.6928\pm 0.0046}$ &
  $0.0196\pm0.0095$ \\ \hline
\multicolumn{1}{|l|}{DCT} &
  \multicolumn{1}{c|}{-} &
  $0.0048\pm0.0084$ &
  $0.1356\pm0.0023$ &
  $0.0312\pm0.0017$ \\ \hline
\multicolumn{1}{|l|}{Vanilla VAE ($z$)} &
  \multicolumn{1}{c|}{-} &
  $0.7090\pm0.0573$ &
  $0.4996\pm0.0265$ &
  $0.6020\pm0.0637$ \\ \hline
\multicolumn{1}{|c|}{\multirow{3}{*}{\begin{tabular}[c]{@{}c@{}}ISVAE\\ vanila dec.\\  ($z$)\end{tabular}}} &
   $4$ &
  $0.7055\pm0.0900$ &
  $0.4660\pm0.0115$ &
  $0.6951\pm0.0934$ \\
\multicolumn{1}{|c|}{} &
  $5$ &
  $0.7131\pm0.0478$ &
  $0.4597\pm0.0029$ &
  $0.6279\pm0.1130$ \\
\multicolumn{1}{|c|}{} &
   $6$ &
  $0.7371\pm0.0718$ &
  $0.5119\pm0.1091$ &
  $\bold{0.7222\pm0.0823}$ \\ \hline
\multicolumn{1}{|c|}{\multirow{3}{*}{\begin{tabular}[c]{@{}c@{}}ISVAE\\ vanila dec.\\ $\boldsymbol{f}_0$\end{tabular}}} &
   &
  $\bold{0.7883\pm0.0853}$ &
  $0.2931\pm0.2491$ &
  $0.5221\pm0.2379$ \\
\multicolumn{1}{|c|}{} &
   &
  $0.7535\pm0.1131$ &
  $0.4259\pm0.1730$ &
  $0.5156\pm0.1032$ \\
\multicolumn{1}{|c|}{} &
   &
  $0.7399\pm0.1504$ &
  $0.4097\pm0.1609$ &
  $0.4530\pm0.1244$ \\ \hline
\multicolumn{1}{|c|}{\multirow{3}{*}{\begin{tabular}[c]{@{}c@{}}ISVAE\\ vanila dec.\\ $\boldsymbol{f}_0$ (ext)\end{tabular}}} &
   &
  $0.7009\pm0.0854$ &
  $0.4836\pm0.0943$ &
  $0.0137\pm0.0217$ \\
\multicolumn{1}{|c|}{} &
   &
  $0.6945\pm0.1459$ &
  $0.5157\pm0.0636$ &
  $0.0055\pm0.0016$ \\
\multicolumn{1}{|c|}{} &
   &
  $0.7252\pm0.0867$ &
  $0.4750\pm0.0873$ &
  $0.0062\pm0.0012$ \\ \hline
\multicolumn{1}{|c|}{\multirow{3}{*}{\begin{tabular}[c]{@{}c@{}}ISVAE\\ attentive dec.\\ ($z$)\end{tabular}}} &
   &
  $0.7117\pm0.0931$ &
  $0.5004\pm0.0942$ &
  $0.6012\pm0.1302$ \\
\multicolumn{1}{|c|}{} &
   &
  $0.7202\pm0.0752$ &
  $0.4958\pm0.0705$ &
  $0.6604\pm0.0427$ \\
\multicolumn{1}{|c|}{} &
   &
  $0.7261\pm0.0652$ &
  $0.4692\pm0.0152$ &
  $0.5946\pm0.0930$ \\ \hline
\multicolumn{1}{|c|}{\multirow{3}{*}{\begin{tabular}[c]{@{}c@{}}ISVAE\\ attentive dec.\\ $\boldsymbol{f}_0$\end{tabular}}} &
   &
  $0.7113\pm0.1432$ &
  $0.1363\pm0.1558$ &
  $0.3173\pm0.2419$ \\
\multicolumn{1}{|c|}{} &
   &
  $0.6831\pm0.0546$ &
  $0.4549\pm0.2031$ &
  $0.3771\pm0.2826$ \\
\multicolumn{1}{|c|}{} &
   &
  $0.7813\pm0.0681$ &
  $0.4910\pm0.0922$ &
  $0.4103\pm0.1774$ \\ \hline
\multicolumn{1}{|c|}{\multirow{3}{*}{\begin{tabular}[c]{@{}c@{}}ISVAE\\ attentive dec.\\ $\boldsymbol{f}_0$ (ext)\end{tabular}}} &
   &
  $0.5847\pm0.1994$ &
  $0.2843\pm0.1148$ &
  $0.1848\pm0.1718$ \\
\multicolumn{1}{|c|}{} &
   &
  $0.6564\pm0.1304$ &
  $0.4725\pm0.1746$ &
  $0.0093\pm0.0031$ \\
\multicolumn{1}{|c|}{} &
   &
  $0.7216\pm0.1071$ &
  $0.5703\pm0.0851$ &
  $0.0125\pm0.0114$ \\ \hline
\end{tabular}%
\end{adjustbox}
\caption{Test results for HAR dataset without transitions using several clustering methods}
\label{tab:clust_HAR_exp}
\end{table}

\section{Conclusions}

In this work, we introduce a novel deep learning model that serves as a proof of concept where learning and interpretability are architecturally aligned.  ISVAE harnesses the capabilities of Variational Autoencoders (VAE) for clustering purposes by learning a new encoding that offers interpretability over the dataset's spectral information by including an attention module at the bottleneck. Indirect interpretability can also be gained by analyzing the encoding development during training, which can be interpreted as a dynamic hierarchical tree, as explained in the synthetic experiment of Section \ref{synth}.\\

Our experiments show that the proposed model outperforms existing methods in terms of clustering accuracy, in spite of using just 1-D signals. Specifically, our model can be applied to a wide range of real-world scenarios involving time signals, with potential application for change point detection in time distributions where accurate clustering is critical. This is possible because ISVAE is not particularly demanding in computation cost and encoding lives in a low dimensional space.\\

Finally, our study contributes to the understanding of the use of VAEs in clustering and suggests new avenues for future research, such as exploring novel architectures that can be added in a modular fashion to the already existing VAE architectures. \\

\section*{Acknowledgements}
We acknowledge the support by the Spanish government MCIN/AEI/10.13039/501100011033/ FEDER, UE, under grant PID2021-123182OB-I00, by Comunidad de Madrid within the ELLIS Unit Madrid framework and also under grant IND2022/TIC-23550.



\bibliographystyle{elsarticle-num-names} 
\bibliography{refs}

\newpage

\section{Appendix}

\subsection{\centering Annex A: Comparison between vanilla and attentive decoder in supervised HAR dataset (without transitions)}\label{annexb}

 Tables \ref{tab:r_HAR_v1}  and \ref{tab:r_HAR_v2} summarizes the current results for vanilla and attentive decoder respectively. We did not employ a validation partition in this instance for the sake of simplicity and demonstrative purposes, addressing the model's generalization capability in the following section.\\

\begin{table}[H]
\begin{adjustbox}{width=1\columnwidth,center}
\begin{tabular}{|c|cccccccccc|}
\hline
 &\multicolumn{10}{c|}{ISVAE with vanilla decoder} \\
 
\hline
Realization &
  \multicolumn{1}{l}{$1$} &
  \multicolumn{1}{l}{$2$} &
  \multicolumn{1}{l}{$3$} &
  \multicolumn{1}{l}{$4$} &
  \multicolumn{1}{l}{$5$} &
  \multicolumn{1}{l}{$6$} &
  \multicolumn{1}{l}{$7$} &
  \multicolumn{1}{l}{$8$} &
  \multicolumn{1}{l}{$9$} &
  \multicolumn{1}{l|}{$10$} \\ \hline
V-score       & $\bold{0.8347}$ & $0.7603$ & $0.8104$ & $0.7941$ & $0.6856$ & $0.7853$ & $0.6857$ & $0.7404$ & $0.6956$ & $0.7859$ \\ 
V-score (ext) & $0.7370$ & $0.8392$ & $0.8138$ & $\bold{0.8446}$ & $0.7060$ & $0.7994$ & $0.6932$ & $0.7915$ & $0.8119$ & $0.7592$ \\ \hline
Realization &
  \multicolumn{1}{l}{$11$} &
  \multicolumn{1}{l}{$12$} &
  \multicolumn{1}{l}{$13$} &
  \multicolumn{1}{l}{$14$} &
  \multicolumn{1}{l}{$15$} &
  \multicolumn{1}{l}{$16$} &
  \multicolumn{1}{l}{$17$} &
  \multicolumn{1}{l}{$18$} &
  \multicolumn{1}{l}{$19$} &
  \multicolumn{1}{l|}{$20$} \\ \hline
V-score       & $0.7172$ & $0.7909$ & $0.6852$ & $0.7736$ & $0.4435$ & $0.6943$ & $0.7597$ & $0.7834$ & $0.7378$ & $0.7625$\\ 
V-score (ext) & $0.7704$ & $0.7641$ & $0.7351$ & $0.8107$ & $0.4912$ & $0.7576$ & $0.7952$ & $0.7596$ & $0.7602$ & $0.8264$ \\ \hline\cline{1-1}
\end{tabular}
\end{adjustbox}
\caption{Clustering results (basic and extended configuration) of each realization over training data. Criteria: highest V-score achieved during training. Mean results V-score $= 0.7326\pm 0.0445$ and V-score (ext) $= 0.7546\pm0.0470$}
\label{tab:r_HAR_v1}
\end{table}

\begin{table}[H]
\begin{adjustbox}{width=1\columnwidth,center}
\begin{tabular}{|c|cccccccccc|}
\hline
 &\multicolumn{10}{c|}{ISVAE with attentive decoder} \\
 
\hline
Realization &
  \multicolumn{1}{l}{$1$} &
  \multicolumn{1}{l}{$2$} &
  \multicolumn{1}{l}{$3$} &
  \multicolumn{1}{l}{$4$} &
  \multicolumn{1}{l}{$5$} &
  \multicolumn{1}{l}{$6$} &
  \multicolumn{1}{l}{$7$} &
  \multicolumn{1}{l}{$8$} &
  \multicolumn{1}{l}{$9$} &
  \multicolumn{1}{l|}{$10$} \\ \hline
V-score       & $0.7527$ & $0.7643$ & $0.7900$ & $0.4587$ & $0.7114$ & $0.6175$ & $0.7707$ & $0.6868$ & $0.6517$ & $0.7759$ \\ 
V-score (ext) & $0.8009$ & $0.8294$ & $0.7270$ & $0.5056$ & $0.7986$ & $0.6772$ & $0.7956$ & $0.7263$ & $0.7314$ & $0.7344$ \\ \hline
Realization &
  \multicolumn{1}{l}{$11$} &
  \multicolumn{1}{l}{$12$} &
  \multicolumn{1}{l}{$13$} &
  \multicolumn{1}{l}{$14$} &
  \multicolumn{1}{l}{$15$} &
  \multicolumn{1}{l}{$16$} &
  \multicolumn{1}{l}{$17$} &
  \multicolumn{1}{l}{$18$} &
  \multicolumn{1}{l}{$19$} &
  \multicolumn{1}{l|}{$20$} \\ \hline
V-score       & $0.6932$ & $\bold{0.8007}$ & $0.7854$ & $0.78711$ & $0.7131$ & $0.7145$ & $0.7284$ & $0.7778$ & $0.7426$ & $0.6816$\\ 
V-score (ext) & $0.7958$ & $0.7374$ & $0.7595$ & $0.8179$ & $0.6707$ & $0.6727$ & $0.7819$ & $\bold{0.8309}$ & $0.7653$ & $0.7226$ \\ \hline\cline{1-1}
\end{tabular}
\end{adjustbox}
\caption{Clustering results (basic and extended configuration) of each realization over training data. Criteria: highest V-score achieved during training. Mean results V-score $= 0.7202\pm 0.0773$ and V-score (ext) $= 0.7440\pm0.0731$ }
\label{tab:r_HAR_v2}
\end{table}

\begin{table}[H]
\begin{adjustbox}{width=0.5\columnwidth,center}
\begin{tabular}{|c|cc|}
 \hline
Decoder       & Vanilla             & Attentive          \\ \hline
V-score       & $\bold{0.7326\pm 0.0445}$ &  $0.7202\pm 0.0773$\\
V-score (ext) & $\bold{0.7546\pm0.0470}$ &  $0.7440\pm0.0731$  \\
\hline
\end{tabular}
\end{adjustbox}
\caption{Summary of results over training data}
\label{tab:summ_HAR}
\end{table}

From a quantitative standpoint and focusing solely on the clustering scores, both decoders yield fairly comparable results, with the standard decoder marginally outperforming the attentive one in both clustering configurations, as outlined in Table \ref{tab:summ_HAR}. Moreover, the maximum V-score - in both basic and extended configurations - are obtained using the vanilla decoder (max V-score $= 0.8347$ and max V-score (ext) $= 0.8446$).\\

Regarding the clustering configurations, generally the extended configuration yields improved results over the basic one. However, it is worth noting that these results represent a composite of two separate yet sequentially linked methods: ISVAE and our selected clustering algorithm, K-means.

\subsection{\centering Annex B: Learned spaces of the vanilla  decoder in supervised HAR dataset (without transitions) (realization 3)}\label{annexc}
 \begin{figure}[H]
\begin{adjustbox}{width=1.5\columnwidth,center}
\begin{tabular}{ccc}
&\includegraphics[width=0.15\textwidth]{images/legend_HAR.pdf} &\\
   \includegraphics[width=0.33\textwidth]{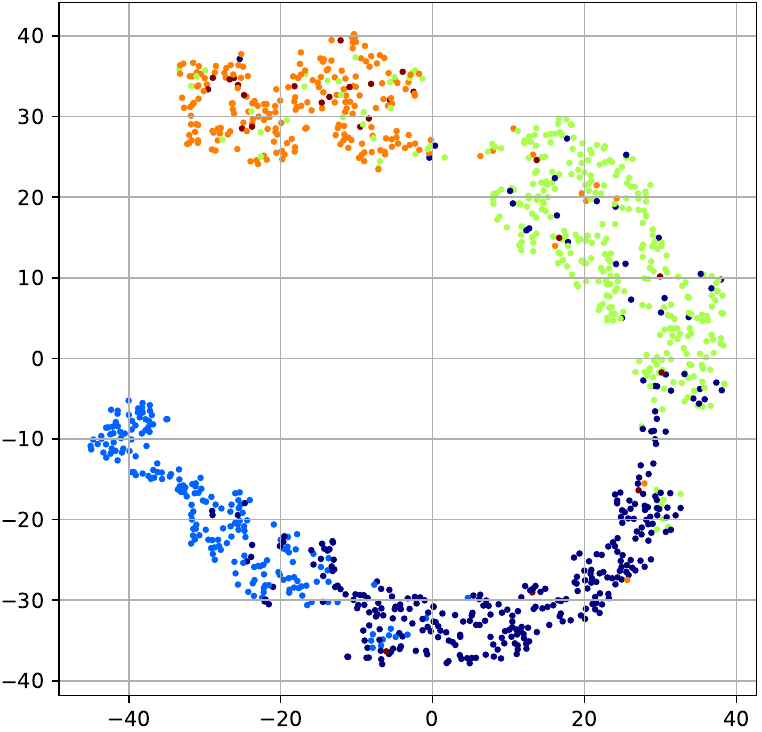} &   \includegraphics[width=0.33\textwidth]{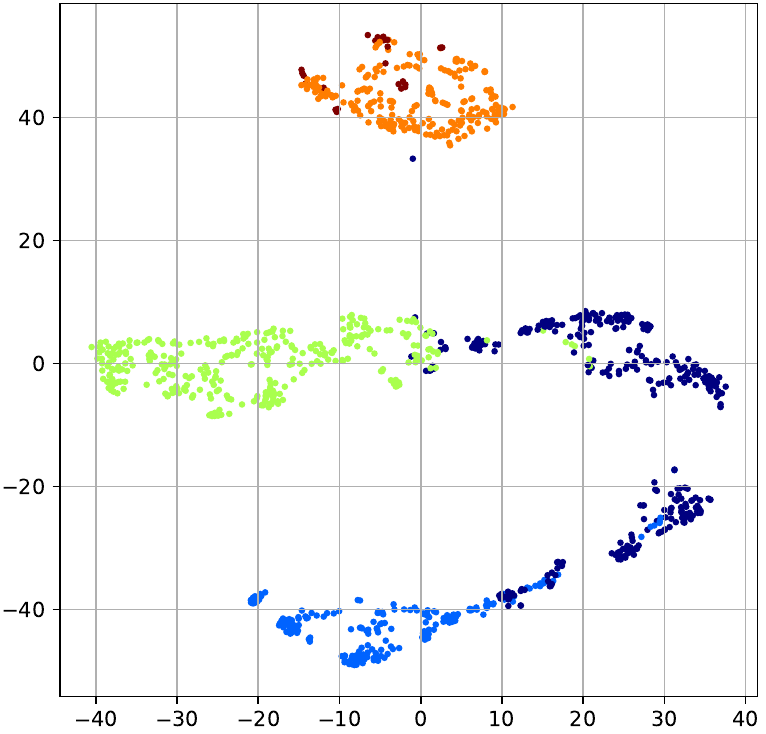}& \includegraphics[width=0.33\textwidth]{images/f_enc_ext3_simple_dec.pdf} \\
   \includegraphics[width=0.33\textwidth]{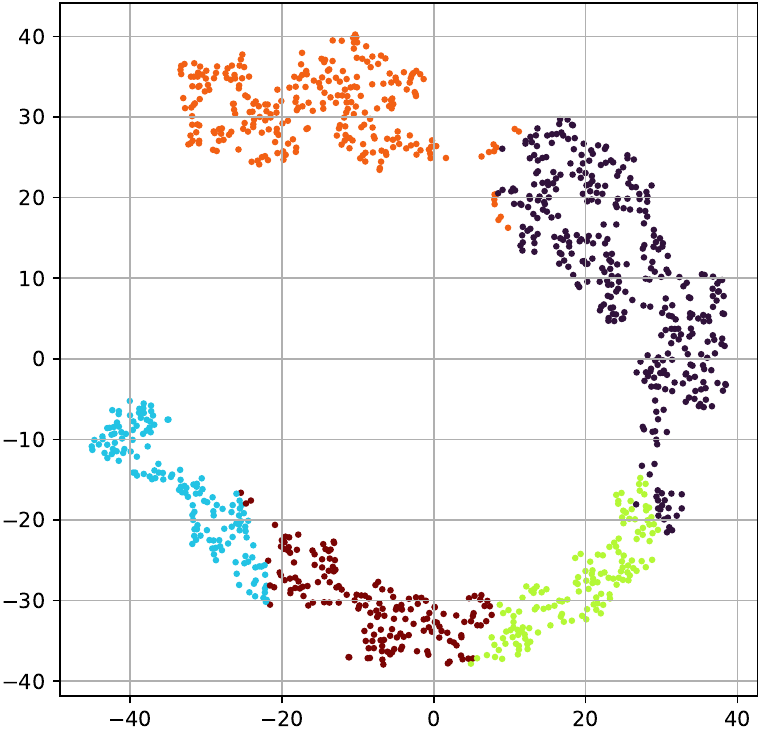} &   \includegraphics[width=0.33\textwidth]{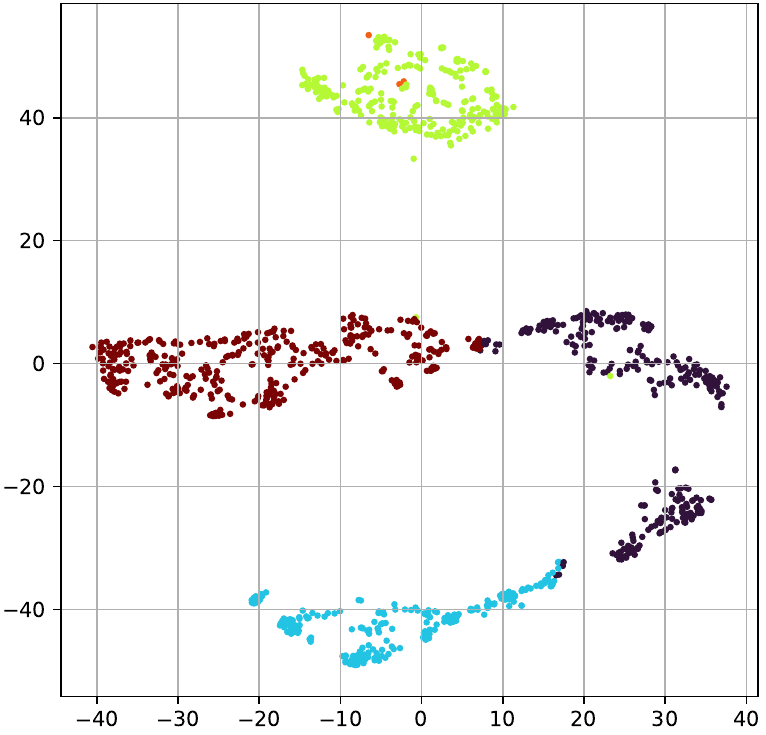}&   \includegraphics[width=0.33\textwidth]{images/f_enc_ext3_KMEANS_simple_dec.pdf} \\
   {\tiny Latent space} & {\tiny$\mathbf{f}_0$  }& {\tiny$\mathbf{f}_0$ (ext.)} \\
   {\tiny V-score $ = 0.6669$ } & {\tiny V-score $ = 0.8104 $ }& {\tiny V-score$ =  0.8138$} \\

\end{tabular}
\end{adjustbox}
\caption{\textbf{ISVAE with vanilla decoder}. First row: Ground truth labels. Second row: K-means predicted labels}
\label{fig:simple_dec_har}
\end{figure}

\subsection{\centering Annex C:  Comparison between vanilla and attentive decoder in unsupervised HAR dataset (with transitions)}\label{annexd}

In order to fairly compare ISVAE's performance (using both decoder versions) in the current dataset against the supervised one, the same procedure has been followed: $20$ realizations each with $300$ epochs, simultaneously run over basic and extended clustering configurations. In terms of hyper-parameter selection, no variation in any of them has been applied. Moreover, $\mathbf{z} \in \mathbb{R}^3$, $J=6$ and $\sigma = 6$. Tables \ref{tab:r_HAR_trans_v1} and \ref{tab:r_HAR_trans_v2} show the results for vanilla and attentive decoder, respectively.

\begin{table}[H]
\begin{adjustbox}{width=1\columnwidth,center}
\begin{tabular}{|c|cccccccccc|}
\hline
 &\multicolumn{10}{c|}{ISVAE with vanilla decoder} \\
 
\hline
Realization &
  \multicolumn{1}{l}{$1$} &
  \multicolumn{1}{l}{$2$} &
  \multicolumn{1}{l}{$3$} &
  \multicolumn{1}{l}{$4$} &
  \multicolumn{1}{l}{$5$} &
  \multicolumn{1}{l}{$6$} &
  \multicolumn{1}{l}{$7$} &
  \multicolumn{1}{l}{$8$} &
  \multicolumn{1}{l}{$9$} &
  \multicolumn{1}{l|}{$10$} \\ \hline
V-score       & $0.6436$ & $0.6034$ & $0.6160$ & $\bold{0.6666}$ & $0.5606$ & $0.6277$ & $0.5735$ & $0.6010$ & $0.5858$ & $0.6123$ \\ 
V-score (ext) & $0.6118$ & $0.6172$ & $0.6553$ & $\bold{0.6798}$ & $0.6282$ & $0.6542$ & $0.6530$ & $0.6109$ & $0.6009$ & $0.6549$ \\ \hline
Realization &
  \multicolumn{1}{l}{$11$} &
  \multicolumn{1}{l}{$12$} &
  \multicolumn{1}{l}{$13$} &
  \multicolumn{1}{l}{$14$} &
  \multicolumn{1}{l}{$15$} &
  \multicolumn{1}{l}{$16$} &
  \multicolumn{1}{l}{$17$} &
  \multicolumn{1}{l}{$18$} &
  \multicolumn{1}{l}{$19$} &
  \multicolumn{1}{l|}{$20$} \\ \hline
V-score       & $0.6194$ & $0.6118$ & $0.5927$ & $0.6209$ & $0.6363$ & $0.6374$ & $0.6428$ & $0.5716$ & $0.5924$ & $0.5971$\\ 
V-score (ext) & $0.6353$ & $0.6542$ & $0.6649$ & $0.6285$ & $0.6588$ & $0.6544$ & $0.6296$ & $0.6087$ & $0.5986$ & $0.5601$ \\ \hline\cline{1-1}
\end{tabular}
\end{adjustbox}
\caption{Clustering results (basic and extended configuration) of each realization over training data. Criteria: highest V-score achieved during training. Mean results V-score $= 0.6106\pm 0.0265$ and V-score (ext) $= 0.6329\pm0.0282$}
\label{tab:r_HAR_trans_v1}
\end{table}

\begin{table}[H]
\begin{adjustbox}{width=1\columnwidth,center}
\begin{tabular}{|c|cccccccccc|}
\hline
 &\multicolumn{10}{c|}{ISVAE with attentive decoder} \\
 
\hline
Realization &
  \multicolumn{1}{l}{$1$} &
  \multicolumn{1}{l}{$2$} &
  \multicolumn{1}{l}{$3$} &
  \multicolumn{1}{l}{$4$} &
  \multicolumn{1}{l}{$5$} &
  \multicolumn{1}{l}{$6$} &
  \multicolumn{1}{l}{$7$} &
  \multicolumn{1}{l}{$8$} &
  \multicolumn{1}{l}{$9$} &
  \multicolumn{1}{l|}{$10$} \\ \hline
V-score       & $\bold{0.6854}$ & $0.6427$ & $0.5991$ & $0.6179$ & $0.5792$ & $0.6010$ & $0.6366$ & $0.5975$ & $0.6000$ & $0.5592$ \\ 
V-score (ext) & $0.6841$ & $\bold{0.6902}$ & $0.6572$ & $0.6509$ & $0.6493$ & $0.6539$ & $0.6358$ & $0.6693$ & $0.6402$ & $0.5648$ \\ \hline
Realization &
  \multicolumn{1}{l}{$11$} &
  \multicolumn{1}{l}{$12$} &
  \multicolumn{1}{l}{$13$} &
  \multicolumn{1}{l}{$14$} &
  \multicolumn{1}{l}{$15$} &
  \multicolumn{1}{l}{$16$} &
  \multicolumn{1}{l}{$17$} &
  \multicolumn{1}{l}{$18$} &
  \multicolumn{1}{l}{$19$} &
  \multicolumn{1}{l}{$20$} \\ \hline
V-score       & $0.6402$ & $0.6145$ & $0.5850$ & $0.6425$ & $0.6102$ & $0.5783$ & $0.6002$ & $0.6503$ & $0.6028$ & $0.6098$\\ 
V-score (ext) & $0.6604$ & $0.6340$ & $0.6357$ & $0.6275$ & $0.6213$ & $0.6169$ & $0.6258$ & $0.6673$ & $0.5986$ & $0.6158$ \\ \hline\cline{1-1}
\end{tabular}
\end{adjustbox}
\caption{Clustering results (basic and extended configuration) of each realization over training data. Criteria: highest V-score achieved during training. Mean results V-score $= 0.6126\pm 0.0289$ and V-score (ext) $= 0.6399\pm0.0286$ }
\label{tab:r_HAR_trans_v2}
\end{table}

\begin{table}[H]
\begin{adjustbox}{width=0.5\columnwidth,center}
\begin{tabular}{|c|cc|}
 \hline
Decoder       & Vanilla             & Attentive          \\ \hline
V-score       & $0.6106\pm 0.0265$ &  $\bold{0.6126\pm 0.0289}$\\
V-score (ext) & $0.6329\pm0.0282$ &  $\bold{0.6399\pm0.0286}$  \\
\hline
\end{tabular}
\end{adjustbox}
\caption{Summary of results over training data}
\label{tab:summ_HAR_trans}
\end{table}

Similar to the supervised HAR dataset and using the quantitative results of the clustering task, differences between decoders are marginal, giving a slightly advantage to the attentive decoder, both in mean results (Table \ref{tab:summ_HAR_trans}) and maximum scores (max V-score $= 0.6854$ and max V-score (ext) $= 0.6902$). Also, the extended configuration reaffirms its superiority in general terms except some exceptions. \\

\subsection{\centering  Annex D: Learned spaces of attentive  decoder in unsupervised HAR dataset (with transitions) (realization 1)}\label{annexe}
\begin{figure}[H]
\begin{adjustbox}{width=1.5\columnwidth,center}
\begin{tabular}{ccc}
    &\includegraphics[width=0.15\textwidth]{images/HAR_trans_legend.pdf} &\\
   \includegraphics[width=0.33\textwidth]{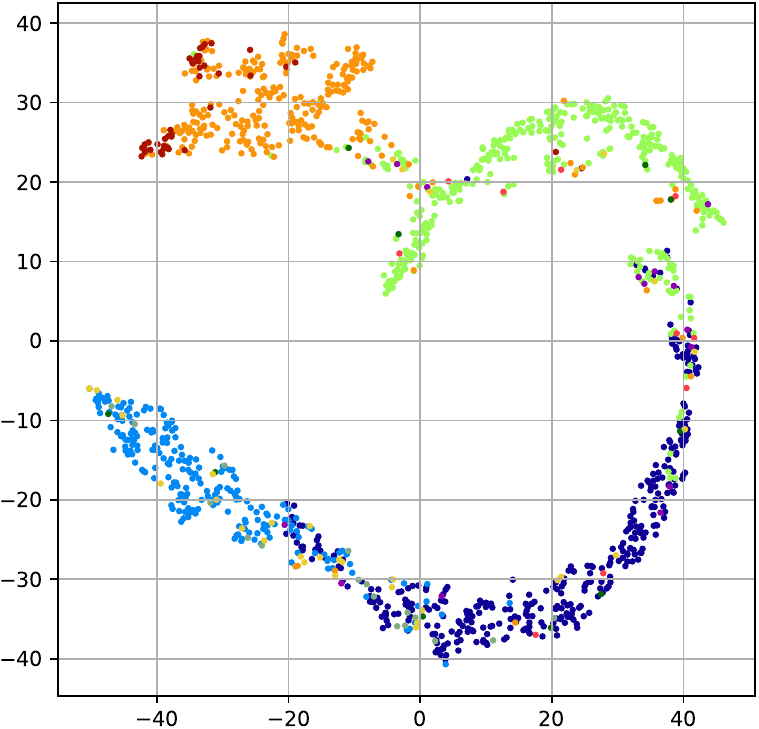} &   \includegraphics[width=0.33\textwidth]{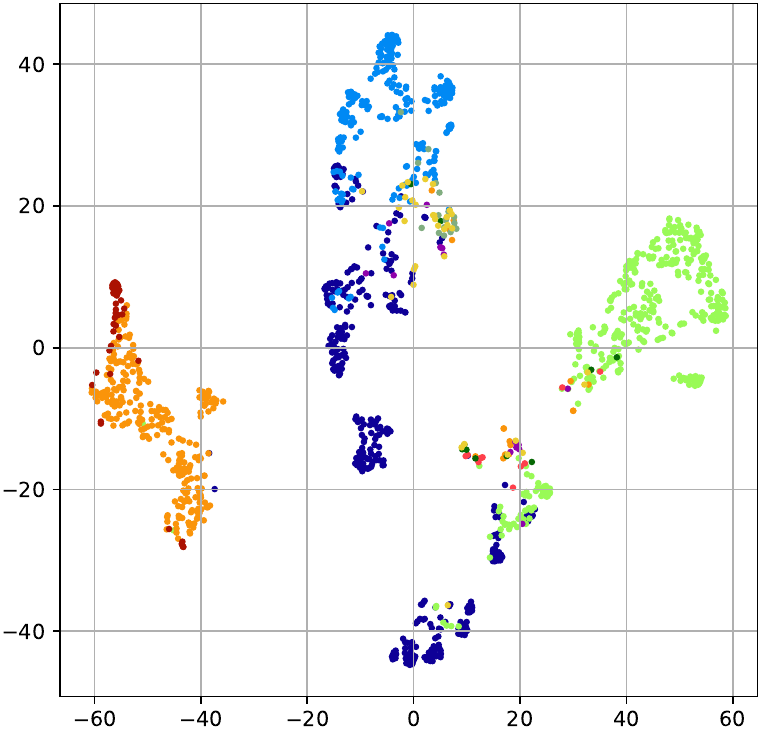}& \includegraphics[width=0.33\textwidth]{images/f_enc_ext3_complex_trans.pdf} \\
   \includegraphics[width=0.33\textwidth]{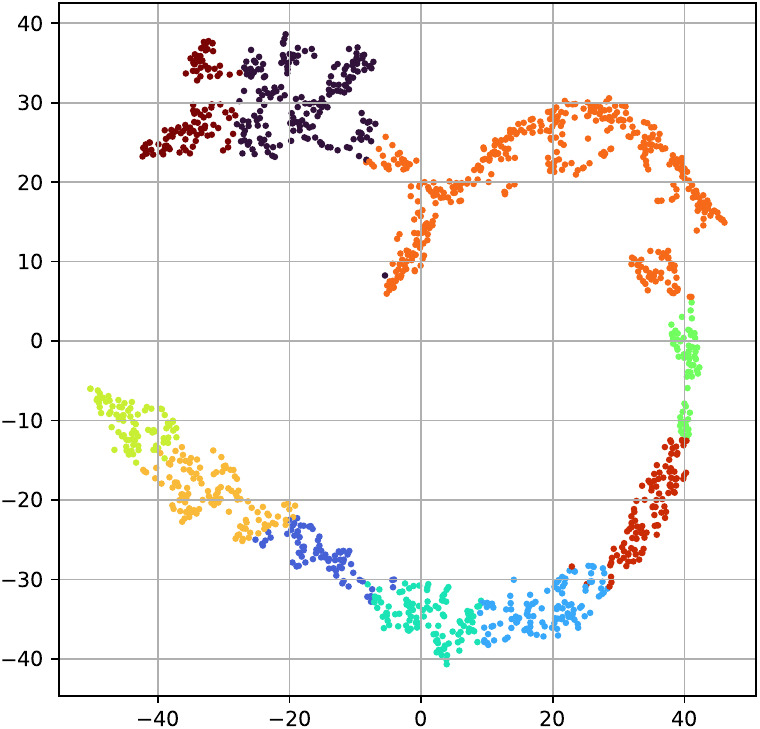} &   \includegraphics[width=0.33\textwidth]{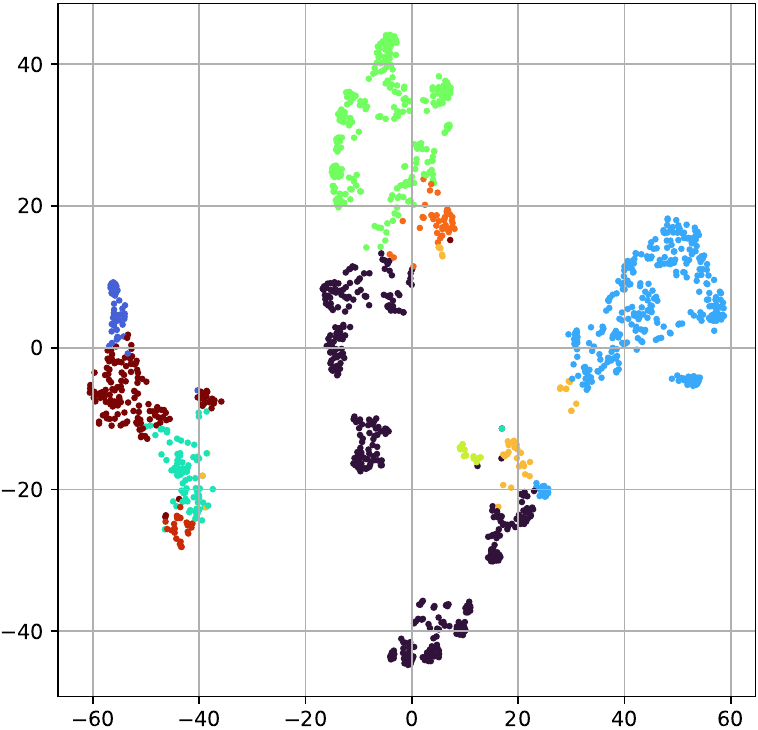}&   \includegraphics[width=0.33\textwidth]{images/f_enc_ext3_kmeans_complex_trans.pdf} \\
   {\tiny Latent space} & {\tiny$\mathbf{f}_0$  }& {\tiny$\mathbf{f}_0$ (ext.)} \\
   {\tiny V-score $ = 0.5808$} & {\tiny V-score $ = 0.6855$ }& {\tiny V-score$ =  0.6842$} \\

\end{tabular}
\end{adjustbox}
\caption{\textbf{ISVAE with attentive decoder}. First row: Ground truth labels. Second row: K-means predicted labels}
\label{fig:complex_dec_har_trans}
\end{figure}

\subsection{\centering  Annex E: Architecture details of ISVAE for each dataset}\label{annexf}
\begin{table}[H]
\begin{adjustbox}{width=1.6\columnwidth,center}
\begin{tabular}{|c|ccc|cc|cc|cc|cc|cc|}
\hline
 &
  \multicolumn{5}{c|}{} &
  \multicolumn{8}{c|}{VAE} \\ \hline
Module &
  \multicolumn{5}{c|}{Filter Bank} &
  \multicolumn{2}{c|}{Encoder} &
  \multicolumn{2}{c|}{\begin{tabular}[c]{@{}c@{}}Decoder\\ (vanilla)\end{tabular}} &
  \multicolumn{4}{c|}{\begin{tabular}[c]{@{}c@{}}Decoder\\ (attentive)\end{tabular}} \\ \hline
NNs &
  \multicolumn{3}{c|}{$\text{CNN}_{\boldsymbol{\eta}_1}$} &
  \multicolumn{2}{c|}{$\text{DNN}_{\boldsymbol{\eta}_2}$} &
  \multicolumn{2}{c|}{$\text{DNN}_{\boldsymbol{\phi}}$} &
  \multicolumn{2}{c|}{$\text{DNN}_{\boldsymbol{\theta}}$} &
  \multicolumn{2}{c|}{$\text{DNN}_{\boldsymbol{\theta}_1}$} &
  \multicolumn{2}{c|}{$\text{DNN}_{\boldsymbol{\theta}_2}$} \\ \hline
 &
  \multicolumn{1}{c|}{Layer Type} &
  \multicolumn{1}{c|}{\begin{tabular}[c]{@{}c@{}}Neurons/filters\\ (stride $=2$)\end{tabular}} &
  \multicolumn{1}{c|}{Activation/regularization} &
  \multicolumn{1}{c|}{Neurons/layer} &
  \multicolumn{1}{c|}{Activation/regularization} &
  \multicolumn{1}{c|}{Neurons/layer} &
  \multicolumn{1}{c|}{Activation/regularization} &
  \multicolumn{1}{c|}{Neurons/layer} &
  \multicolumn{1}{c|}{Activation/regularization} &
  \multicolumn{1}{c|}{Neurons/layer} &
  \multicolumn{1}{c|}{Activation/regularization} &
  \multicolumn{1}{c|}{Neurons/layer} &
  \multicolumn{1}{c|}{Activation/regularization} \\ \hline
\multirow{3}{*}{Synthetic} &
  Input &
  $3\times3$ &
  Maxpool (1D) ($3\times3$, stride = 2) &
  36 &
  ReLU &
  $J$ &
  ReLU &
  $K$ &
  ReLU &
  $K$ &
  ReLU &
  $J+K$ &
  ReLU \\
 &
  Hidden &
  - &
  - &
  {[}20,10{]} &
  ReLU &
  $J$ &
  ReLU &
  {[}70,150,300{]} &
  ReLU &
  {[}$J$,$J${]} &
  ReLU &
  {[}70,150,300{]} &
  ReLU \\
 &
  Output &
  $3\times3$ &
  ReLU &
  1 &
  Sigmoid &
  $J$ &
  - &
  $D=600$ &
  - &
  $J$ &
  Sigmoid &
  600 &
  - \\ \hline
\multirow{3}{*}{\begin{tabular}[c]{@{}c@{}}HAR\\ (with \& without\\ transitions)\end{tabular}} &
  Input &
  $3\times3$ &
  Maxpool (1D) ($3\times3$, stride = 2) &
  6 &
  ReLU &
  $J$ &
  ReLU &
  $K$ &
  ReLU &
  $K$ &
  ReLU &
  $J+K$ &
  ReLU \\
 &
  Hidden &
  - &
  - &
  5 &
  ReLU &
  $J$ &
  ReLU &
  50 &
  ReLU &
  {[}$J$,$J${]} &
  ReLU &
  50 &
  ReLU \\
 &
  Output &
  $3\times3$ &
  ReLU &
  1 &
  Sigmoid &
  $J$ &
  - &
  $D=112$ &
  - &
  $J$ &
  Sigmoid &
  $D=112$ &
  - \\ \hline
\multirow{3}{*}{Active HAR} &
  Input &
  $3\times3$ &
  Maxpool (1D) ($3\times3$, stride = 2) &
  6 &
  ReLU &
  $J$ &
  ReLU &
  $K$ &
  ReLU &
  $K$ &
  ReLU &
  $J+K$ &
  ReLU \\
 &
  Hidden &
  - &
  - &
  5 &
  ReLU &
  $J$ &
  ReLU &
  {[}20,40,80{]} &
  ReLU &
  {[}$J$,$J${]} &
  ReLU &
  {[}20,40,80{]} &
  ReLU \\
 &
  Output &
  $3\times3$ &
  ReLU &
  1 &
  Sigmoid &
  $J$ &
  - &
  $D=125$ &
  - &
  $J$ &
  Sigmoid &
  $D=125$ &
  - \\ \hline
\multirow{3}{*}{\begin{tabular}[c]{@{}c@{}}SODA\\ (accelerometer \&\\ gyroscope)\end{tabular}} &
  Input &
  $3\times3$ &
  Maxpool (1D) ($3\times3$, stride = 2) &
  5 &
  ReLU &
  $J$ &
  ReLU &
  $K$ &
  ReLU &
  $K$ &
  ReLU &
  $J+K$ &
  ReLU \\
 &
  Hidden &
  - &
  - &
  5 &
  ReLU &
  $J$ &
  ReLU &
  {[}20,40,80{]} &
  ReLU &
  {[}$J$,$J${]} &
  ReLU &
  {[}20,40,80{]} &
  ReLU \\
 &
  Output &
  $3\times3$ &
  ReLU &
  1 &
  Sigmoid &
  $J$ &
  - &
  $D=100$ &
  - &
  $J$ &
  Sigmoid &
  $D=100$ &
  - \\ \hline
\end{tabular}
\end{adjustbox}
\caption{ISVAE's networks architecture for each dataset}
\end{table}

\subsection{\centering  Annex F: Performance comparison between clustering methods over several datasets}\label{annexg}

\begin{table}[H]
\begin{adjustbox}{width=0.8\columnwidth,center}
\begin{tabular}{cc|ccc|}
\cline{3-5}
\multicolumn{1}{l}{\multirow{2}{*}{}} &
  \multicolumn{1}{l|}{\multirow{2}{*}{}} &
  \multicolumn{3}{c|}{\multirow{2}{*}{Dataset \#2: HAR (transitions)}} \\
\multicolumn{1}{l}{}       & \multicolumn{1}{l|}{} & \multicolumn{3}{c|}{}                                     \\ \cline{2-5} 
\multicolumn{1}{l|}{}      & \multicolumn{1}{l|}{} & \multicolumn{3}{c|}{V-score}                              \\ \cline{2-5} 
\multicolumn{1}{l|}{} &
  \multicolumn{1}{l|}{J} &
  \multicolumn{1}{c|}{K-Means} &
  \multicolumn{1}{c|}{DBSCAN} &
  \multicolumn{1}{c|}{Spectral} \\ \hline
\multicolumn{1}{|l|}{Time} & \multicolumn{1}{c|}{-} & $0.6048\pm0.0078$ & $\bold{0.6231\pm0.0021}$ & $0.0251\pm0.0016$ \\ \hline
\multicolumn{1}{|l|}{DCT}  & \multicolumn{1}{c|}{-} & $0.0023\pm0.0056$ & $0.1715\pm0.0038$ & $0.0114\pm0.0072$ \\ \hline
\multicolumn{1}{|l|}{Vanilla VAE ($z$)} &
  \multicolumn{1}{c|}{-} &
  $0.5602\pm0.0155$ &
  $0.4271\pm0.0064$ &
  $\bold{0.5866\pm0.0276}$ \\ \hline
\multicolumn{1}{|c|}{\multirow{3}{*}{\begin{tabular}[c]{@{}c@{}}ISVAE\\ vanila dec.\\  ($z$)\end{tabular}}} &
  $4$ &
  $0.5494\pm0.0089$ &
  $0.4328\pm0.0162$ &
  $0.5358\pm0.0553$ \\
\multicolumn{1}{|c|}{}     & $5$                   & $0.5402\pm0.0323$ & $0.4304\pm0.0103$ & $0.5100\pm0.0395$ \\
\multicolumn{1}{|c|}{}     & $6$                   & $0.4631\pm0.1773$ & $0.2746\pm0.1832$ & $0.4512\pm0.1792$ \\ \hline
\multicolumn{1}{|c|}{\multirow{3}{*}{\begin{tabular}[c]{@{}c@{}}ISVAE\\ vanila dec.\\ $\boldsymbol{f}_0$\end{tabular}}} &
   &
  $0.7009\pm0.0655$ &
  $0.2892\pm0.1641$ &
  $0.3174\pm0.1363$ \\
\multicolumn{1}{|c|}{}     &                       & $0.6864\pm0.1121$ & $0.2423\pm0.1689$ & $0.1002\pm0.0765$ \\
\multicolumn{1}{|c|}{}     &                       & $0.6859\pm0.0929$ & $0.2643\pm0.1251$ & $0.2886\pm0.1404$ \\ \hline
\multicolumn{1}{|c|}{\multirow{3}{*}{\begin{tabular}[c]{@{}c@{}}ISVAE\\ vanila dec.\\ $\boldsymbol{f}_0$ (ext)\end{tabular}}} &
   &
  $0.7112\pm0.0662$ &
  $0.4010\pm0.0918$ &
  $0.0813\pm0.1389$ \\
\multicolumn{1}{|c|}{}     &                       & $0.6926\pm0.0787$ & $0.3651\pm0.1157$ & $0.0045\pm0.0021$ \\
\multicolumn{1}{|c|}{}     &                       & $\bold{0.7116\pm0.0740}$ & $0.3585\pm0.0507$ & $0.0064\pm0.0048$ \\ \hline
\multicolumn{1}{|c|}{\multirow{3}{*}{\begin{tabular}[c]{@{}c@{}}ISVAE\\ attentive dec.\\ ($z$)\end{tabular}}} &
   &
  $0.4656\pm0.1753$ &
  $0.3425\pm0.1489$ &
  $0.4486\pm0.1740$ \\
\multicolumn{1}{|c|}{}     &                       & $0.5983\pm0.0514$ & $0.4579\pm0.0791$ & $0.6102\pm0.0634$ \\
\multicolumn{1}{|c|}{}     &                       & $0.6347\pm0.0632$ & $0.4153\pm0.0115$ & $0.6423\pm0.0835$ \\ \hline
\multicolumn{1}{|c|}{\multirow{3}{*}{\begin{tabular}[c]{@{}c@{}}ISVAE\\ attentive dec.\\ $\boldsymbol{f}_0$\end{tabular}}} &
   &
  $0.6859\pm0.0490$ &
  $0.1677\pm0.1004$ &
  $0.1532\pm0.1497$ \\
\multicolumn{1}{|c|}{}     &                       & $0.6013\pm0.1210$ & $0.1375\pm0.1597$ & $0.1121\pm0.0699$ \\
\multicolumn{1}{|c|}{}     &                       & $0.6695\pm0.0628$ & $0.3961\pm0.1423$ & $0.2081\pm0.0874$ \\ \hline
\multicolumn{1}{|c|}{\multirow{3}{*}{\begin{tabular}[c]{@{}c@{}}ISVAE\\ attentive dec.\\ $\boldsymbol{f}_0$ (ext)\end{tabular}}} &
   &
  $0.6632\pm0.0639$ &
  $0.3475\pm0.0230$ &
  $0.0954\pm0.1301$ \\
\multicolumn{1}{|c|}{}     &                       & $0.6131\pm0.1301$ & $0.2514\pm0.1263$ & $0.0048\pm0.0016$ \\
\multicolumn{1}{|c|}{}     &                       & $0.6729\pm0.1189$ & $0.4444\pm0.0845$ & $0.0070\pm0.0036$ \\ \hline
\end{tabular}%
\end{adjustbox}
\caption{Test results for HAR (with transitions) dataset without transitions using several clustering methods}
\label{tab:clust_HAR_trans_exp}
\end{table}

\begin{table}[H]
\begin{adjustbox}{width=0.8\columnwidth,center}
\begin{tabular}{cc|ccc|}
\cline{3-5}
\multicolumn{1}{l}{\multirow{2}{*}{}} &
  \multicolumn{1}{l|}{\multirow{2}{*}{}} &
  \multicolumn{3}{c|}{\multirow{2}{*}{Dataset \#3: Active HAR}} \\
\multicolumn{1}{l}{} &
  \multicolumn{1}{l|}{} &
  \multicolumn{3}{c|}{} \\ \cline{2-5} 
\multicolumn{1}{l|}{} &
  \multicolumn{1}{l|}{} &
  \multicolumn{3}{c|}{V-score} \\ \cline{2-5} 
\multicolumn{1}{l|}{} &
  \multicolumn{1}{l|}{J} &
  \multicolumn{1}{c|}{K-Means} &
  \multicolumn{1}{c|}{DBSCAN} &
  \multicolumn{1}{c|}{Spectral} \\ \hline
\multicolumn{1}{|l|}{Time} &
  \multicolumn{1}{c|}{-} &
  $0.4765\pm0.0323$ &
  $0.3394\pm0.0241$ &
  $0.2643\pm0.0483$ \\ \hline
\multicolumn{1}{|l|}{DCT} &
  \multicolumn{1}{c|}{-} &
  $0.2342\pm0.0451$ &
  $0.3462\pm0.0547$ &
  $0.0131\pm0.0409$ \\ \hline
\multicolumn{1}{|l|}{Vanilla VAE ($z$)} &
  \multicolumn{1}{c|}{-} &
  $0.2604\pm0.0987$ &
  $0.0670\pm0.0405$ &
  $0.2548\pm0.0908$ \\ \hline
\multicolumn{1}{|c|}{\multirow{3}{*}{\begin{tabular}[c]{@{}c@{}}ISVAE\\ vanila dec.\\  ($z$)\end{tabular}}} &
  $4$ &
  $0.4823\pm0.1623$ &
  $0.4083\pm0.1498$ &
  $0.4972\pm0.1334$ \\
\multicolumn{1}{|c|}{} &
  $5$ &
  $0.5688\pm0.0861$ &
  $\bold{0.5181\pm0.0588}$ &
  $\bold{0.5785\pm0.0638}$ \\
\multicolumn{1}{|c|}{} &
  $6$ &
  $0.4424\pm0.0807$ &
  $0.3674\pm0.1496$ &
  $0.4448\pm0.1363$ \\ \hline
\multicolumn{1}{|c|}{\multirow{3}{*}{\begin{tabular}[c]{@{}c@{}}ISVAE\\ vanila dec.\\ $\boldsymbol{f}_0$\end{tabular}}} &
   &
  $0.5647\pm0.0756$ &
  $0.3060\pm0.1801$ &
  $0.3010\pm0.1750$ \\
\multicolumn{1}{|c|}{} &
   &
  $0.5567\pm0.0552$ &
  $0.4208\pm0.0287$ &
  $0.4863\pm0.0392$ \\
\multicolumn{1}{|c|}{} &
   &
  $0.5211\pm0.0291$ &
  $0.4126\pm0.0347$ &
  $0.3098\pm0.1765$ \\ \hline
\multicolumn{1}{|c|}{\multirow{3}{*}{\begin{tabular}[c]{@{}c@{}}ISVAE\\ vanila dec.\\ $\boldsymbol{f}_0$ (ext)\end{tabular}}} &
   &
  $0.6311\pm0.1005$ &
  $0.4864\pm0.0525$ &
  $0.2292\pm0.1594$ \\
\multicolumn{1}{|c|}{} &
   &
  $\bold{0.7538\pm0.0607}$ &
  $0.4483\pm0.0612$ &
  $0.1366\pm0.1297$ \\
\multicolumn{1}{|c|}{} &
   &
  $0.7236\pm0.0858$ &
  $0.4812\pm0.0672$ &
  $0.0701\pm0.0622$ \\ \hline
\multicolumn{1}{|c|}{\multirow{3}{*}{\begin{tabular}[c]{@{}c@{}}ISVAE\\ attentive dec.\\ ($z$)\end{tabular}}} &
   &
  $0.2517\pm0.2155$ &
  $0.2322\pm0.2478$ &
  $0.2485\pm0.1959$ \\
\multicolumn{1}{|c|}{} &
   &
  $0.5412\pm0.1470$ &
  $0.4577\pm0.1621$ &
  $0.5469\pm0.1226$ \\
\multicolumn{1}{|c|}{} &
   &
  $0.2940\pm0.2090$ &
  $0.2143\pm0.2172$ &
  $0.2850\pm0.2086$ \\ \hline
\multicolumn{1}{|c|}{\multirow{3}{*}{\begin{tabular}[c]{@{}c@{}}ISVAE\\ attentive dec.\\ $\boldsymbol{f}_0$\end{tabular}}} &
   &
  $0.4844\pm0.1390$ &
  $0.4050\pm0.1466$ &
  $0.1767\pm0.1929$ \\
\multicolumn{1}{|c|}{} &
   &
  $0.6336\pm0.0822$ &
  $0.4273\pm0.0232$ &
  $0.3823\pm0.2333$ \\
\multicolumn{1}{|c|}{} &
   &
  $0.5514\pm0.0627$ &
  $0.4499\pm0.0614$ &
  $0.2642\pm0.1647$ \\ \hline
\multicolumn{1}{|c|}{\multirow{3}{*}{\begin{tabular}[c]{@{}c@{}}ISVAE\\ attentive dec.\\ $\boldsymbol{f}_0$ (ext)\end{tabular}}} &
   &
  $0.6604\pm0.0688$ &
  $0.4897\pm0.0546$ &
  $0.0530\pm0.0648$ \\
\multicolumn{1}{|c|}{} &
   &
  $0.7126\pm0.0921$ &
  $0.4898\pm0.0799$ &
  $0.1679\pm0.1395$ \\
\multicolumn{1}{|c|}{} &
   &
  $0.7086\pm0.0814$ &
  $0.4880\pm0.0334$ &
  $0.1678\pm0.1250$ \\ \hline
\end{tabular}%
\end{adjustbox}
\caption{Test results for active HAR dataset without transitions using several clustering methods}
\label{tab:clust_HAR_act_exp}
\end{table}

\begin{table}[H]
\begin{adjustbox}{width=0.8\columnwidth,center}
\begin{tabular}{cc|ccc|}
\cline{3-5}
\multicolumn{1}{l}{\multirow{2}{*}{}} &
  \multicolumn{1}{l|}{\multirow{2}{*}{}} &
  \multicolumn{3}{c|}{\multirow{2}{*}{Dataset \#4: SoDA (accelerometer)}} \\
\multicolumn{1}{l}{}                    & \multicolumn{1}{l|}{}  & \multicolumn{3}{c|}{}                                                                      \\ \cline{2-5} 
\multicolumn{1}{l|}{}                   & \multicolumn{1}{l|}{}  & \multicolumn{3}{c|}{V-score}                                                               \\ \cline{2-5} 
\multicolumn{1}{l|}{}                   & \multicolumn{1}{l|}{J} & \multicolumn{1}{c|}{K-Means} & \multicolumn{1}{c|}{DBSCAN} & \multicolumn{1}{c|}{Spectral} \\ \hline
\multicolumn{1}{|l|}{Time}              & \multicolumn{1}{c|}{-}  & $0.7012\pm0.0220$            & $0.4112\pm0.0059$           & $0.0477\pm0.0267$             \\ \hline
\multicolumn{1}{|l|}{DCT}               & \multicolumn{1}{c|}{-}  & $0.0276\pm0.0683$            & $0.0751\pm0.1699$           & $0.0190\pm0.0066$             \\ \hline
\multicolumn{1}{|l|}{Vanilla VAE ($z$)} & \multicolumn{1}{c|}{-}  & $0.6583\pm0.0698$            & $\bold{0.6517\pm0.1720}$    & $\bold{0.6558\pm0.0673}$      \\ \hline
\multicolumn{1}{|c|}{\multirow{3}{*}{\begin{tabular}[c]{@{}c@{}}ISVAE\\ vanila dec.\\  ($z$)\end{tabular}}} &
  $4$ &
  $0.6959\pm0.0490$ &
  $0.5021\pm0.2278$ &
  $0.6605\pm0.0694$ \\
\multicolumn{1}{|c|}{}                  & $5$                    & $0.5500\pm0.0952$            & $0.4887\pm0.1111$           & $0.5499\pm0.0886$             \\
\multicolumn{1}{|c|}{}                  & $6$                    & $0.5986\pm0.0510$            & $0.4394\pm0.1691$           & $0.6083\pm0.0451$             \\ \hline
\multicolumn{1}{|c|}{\multirow{3}{*}{\begin{tabular}[c]{@{}c@{}}ISVAE\\ vanila dec.\\ $\boldsymbol{f}_0$\end{tabular}}} &
   &
  $0.6256\pm0.0685$ &
  $0.3094\pm0.2185$ &
  $0.2543\pm0.1390$ \\
\multicolumn{1}{|c|}{}                  &                        & $0.6610\pm0.0855$            & $0.4370\pm0.0536$           & $0.2611\pm0.1564$             \\
\multicolumn{1}{|c|}{}                  &                        & $0.6149\pm0.0960$            & $0.5067\pm0.0654$           & $0.1970\pm0.0929$             \\ \hline
\multicolumn{1}{|c|}{\multirow{3}{*}{\begin{tabular}[c]{@{}c@{}}ISVAE\\ vanila dec.\\ $\boldsymbol{f}_0$ (ext)\end{tabular}}} &
   &
  $0.6415\pm0.1148$ &
  $0.5045\pm0.0864$ &
  $0.2233\pm0.2334$ \\
\multicolumn{1}{|c|}{}                  &                        & $\bold{0.7094\pm0.0899}$     & $0.5555\pm0.0261$           & $0.1004\pm0.1431$             \\
\multicolumn{1}{|c|}{}                  &                        & $0.5770\pm0.0628$            & $0.5469\pm0.0411$           & $0.0160\pm0.0121$             \\ \hline
\multicolumn{1}{|c|}{\multirow{3}{*}{\begin{tabular}[c]{@{}c@{}}ISVAE\\ attentive dec.\\ ($z$)\end{tabular}}} &
   &
  $0.4087\pm0.2507$ &
  $0.2720\pm0.2660$ &
  $0.4175\pm0.2338$ \\
\multicolumn{1}{|c|}{}                  &                        & $0.5029\pm0.2205$            & $0.4370\pm0.2302$           & $0.4710\pm0.2071$             \\
\multicolumn{1}{|c|}{}                  &                        & $0.6498\pm0.0909$            & $0.4431\pm0.0922$           & $0.6383\pm0.0773$             \\ \hline
\multicolumn{1}{|c|}{\multirow{3}{*}{\begin{tabular}[c]{@{}c@{}}ISVAE\\ attentive dec.\\ $\boldsymbol{f}_0$\end{tabular}}} &
   &
  $0.6114\pm0.1001$ &
  $0.3720\pm0.1946$ &
  $0.2906\pm0.2267$ \\
\multicolumn{1}{|c|}{}                  &                        & $0.6002\pm0.1045$            & $0.3541\pm0.1720$           & $0.3237\pm0.2448$             \\
\multicolumn{1}{|c|}{}                  &                        & $0.7095\pm0.0660$            & $0.5053\pm0.0787$           & $0.2960\pm0.2538$             \\ \hline
\multicolumn{1}{|c|}{\multirow{3}{*}{\begin{tabular}[c]{@{}c@{}}ISVAE\\ attentive dec.\\ $\boldsymbol{f}_0$ (ext)\end{tabular}}} &
   &
  $0.5683\pm0.1385$ &
  $0.5012\pm0.1289$ &
  $0.0396\pm0.0298$ \\
\multicolumn{1}{|c|}{}                  &                        & $0.6355\pm0.1182$            & $0.5304\pm0.1232$           & $0.0775\pm0.0951$             \\
\multicolumn{1}{|c|}{}                  &                        & $0.6580\pm0.1034$            & $0.5390\pm0.0339$           & $0.0189\pm0.0178$             \\ \hline
\end{tabular}%
\end{adjustbox}
\caption{Test results for SoDA (accelerometer) dataset without transitions using several clustering methods}
\label{tab:clust_SODA_acc_exp}
\end{table}

\begin{table}[H]
\begin{adjustbox}{width=0.8\columnwidth,center}
\begin{tabular}{cc|ccc|}
\cline{3-5}
\multicolumn{1}{l}{\multirow{2}{*}{}} &
  \multicolumn{1}{l|}{\multirow{2}{*}{}} &
  \multicolumn{3}{c|}{\multirow{2}{*}{Dataset \#5: SoDA (gyroscope)}} \\
\multicolumn{1}{l}{}       & \multicolumn{1}{l|}{} & \multicolumn{3}{c|}{}                                     \\ \cline{2-5} 
\multicolumn{1}{l|}{}      & \multicolumn{1}{l|}{} & \multicolumn{3}{c|}{V-score}                              \\ \cline{2-5} 
\multicolumn{1}{l|}{} &
  \multicolumn{1}{l|}{J} &
  \multicolumn{1}{c|}{K-Means} &
  \multicolumn{1}{c|}{DBSCAN} &
  \multicolumn{1}{c|}{Spectral} \\ \hline
\multicolumn{1}{|l|}{Time} & \multicolumn{1}{c|}{-} & $0.2526\pm0.0404$ & $0.0476\pm0.1200$ & $0.0142\pm0.0073$ \\ \hline
\multicolumn{1}{|l|}{DCT}  & \multicolumn{1}{c|}{-} & $0.0116\pm0.0227$ & $0.0067\pm0.0142$ & $0.0142\pm0.0073$ \\ \hline
\multicolumn{1}{|l|}{Vanilla VAE ($z$)} &
  \multicolumn{1}{c|}{-} &
  $0.3516\pm0.0376$ &
  $0.2078\pm0.1239$ &
  $0.2117\pm0.0661$ \\ \hline
\multicolumn{1}{|c|}{\multirow{3}{*}{\begin{tabular}[c]{@{}c@{}}ISVAE\\ vanila dec.\\  ($z$)\end{tabular}}} &
  $4$ &
  $0.2626\pm0.0448$ &
  $0.1485\pm0.0655$ &
  $0.2519\pm0.0449$ \\
\multicolumn{1}{|c|}{}     & $5$                   & $0.2365\pm0.0413$ & $0.1487\pm0.0894$ & $0.2088\pm0.0543$ \\
\multicolumn{1}{|c|}{}     & $6$                   & $0.1782\pm0.0237$ & $0.1183\pm0.0442$ & $0.1713\pm0.0141$ \\ \hline
\multicolumn{1}{|c|}{\multirow{3}{*}{\begin{tabular}[c]{@{}c@{}}ISVAE\\ vanila dec.\\ $\boldsymbol{f}_0$\end{tabular}}} &
   &
  $0.3504\pm0.0605$ &
  $0.2070\pm0.0858$ &
  $0.1815\pm0.1046$ \\
\multicolumn{1}{|c|}{}     &                       & $0.2851\pm0.1179$ & $0.1203\pm0.0517$ & $0.1247\pm0.0759$ \\
\multicolumn{1}{|c|}{}     &                       & $0.2266\pm0.0672$ & $0.1195\pm0.0578$ & $0.0638\pm0.0402$ \\ \hline
\multicolumn{1}{|c|}{\multirow{3}{*}{\begin{tabular}[c]{@{}c@{}}ISVAE\\ vanila dec.\\ $\boldsymbol{f}_0$ (ext)\end{tabular}}} &
   &
  $\bold{0.4706\pm0.0476}$ &
  $\bold{0.2636\pm0.0681}$ &
  $0.1146\pm0.1185$ \\
\multicolumn{1}{|c|}{}     &                       & $0.3494\pm0.0871$ & $0.1631\pm0.0410$ & $0.0463\pm0.0329$ \\
\multicolumn{1}{|c|}{}     &                       & $0.2977\pm0.0488$ & $0.1162\pm0.0453$ & $0.0245\pm0.0247$ \\ \hline
\multicolumn{1}{|c|}{\multirow{3}{*}{\begin{tabular}[c]{@{}c@{}}ISVAE\\ attentive dec.\\ ($z$)\end{tabular}}} &
   &
  $0.2377\pm0.0743$ &
  $0.1285\pm0.0891$ &
  $0.2182\pm0.0746$ \\
\multicolumn{1}{|c|}{}     &                       & $0.2695\pm0.0657$ & $0.1449\pm0.0863$ & $\bold{0.2518\pm0.0698}$ \\
\multicolumn{1}{|c|}{}     &                       & $0.2503\pm0.0596$ & $0.1977\pm0.0817$ & $0.2342\pm0.0584$ \\ \hline
\multicolumn{1}{|c|}{\multirow{3}{*}{\begin{tabular}[c]{@{}c@{}}ISVAE\\ attentive dec.\\ $\boldsymbol{f}_0$\end{tabular}}} &
   &
  $0.2683\pm0.0530$ &
  $0.1242\pm0.0344$ &
  $0.1150\pm0.1150$ \\
\multicolumn{1}{|c|}{}     &                       & $0.2843\pm0.0580$ & $0.1640\pm0.0567$ & $0.0804\pm0.0655$ \\
\multicolumn{1}{|c|}{}     &                       & $0.2744\pm0.0896$ & $0.1383\pm0.0295$ & $0.0655\pm0.0452$ \\ \hline
\multicolumn{1}{|c|}{\multirow{3}{*}{\begin{tabular}[c]{@{}c@{}}ISVAE\\ attentive dec.\\ $\boldsymbol{f}_0$ (ext)\end{tabular}}} &
   &
  $0.3367\pm0.0601$ &
  $0.1390\pm0.0558$ &
  $0.0628\pm0.0453$ \\
\multicolumn{1}{|c|}{}     &                       & $0.3537\pm0.0683$ & $0.2101\pm0.0562$ & $0.0151\pm0.0180$ \\
\multicolumn{1}{|c|}{}     &                       & $0.3581\pm0.1171$ & $0.1332\pm0.0486$ & $0.0160\pm0.0139$ \\ \hline
\end{tabular}%
\end{adjustbox}
\caption{Test results for SoDA (gyroscope) dataset without transitions using several clustering methods}
\label{tab:clust_SODA_gyr_exp}
\end{table}





\end{document}